\documentclass[10pt,journal,compsoc]{IEEEtran}

\usepackage{listings}
\usepackage{xcolor}
\usepackage{colortbl}
\usepackage{amsfonts,amssymb,amsmath,amsthm}
\usepackage{microtype}
\usepackage{graphicx}
\usepackage{multirow}
\usepackage{array}
\usepackage{subfig}
\usepackage{booktabs}
\usepackage{hyperref}
\usepackage{bm}
\usepackage{multirow}
\usepackage{array}
\usepackage{chemarrow}
\usepackage{arydshln}
\usepackage[noend]{algpseudocode}
\usepackage{algorithmicx,algorithm}
\newtheorem{proposition}{\bf {\em Proposition}}
\newtheorem*{proofA}{\bf {\em Proof}}

\newcommand{\etal}{\textit{et al}. }
\newcounter{MYtempeqncnt}

\newcommand{\eg}{\textit{e.g}.}
\newcommand{\ie}{\textit{i.e}.}
\newcommand{\etc}{\textit{etc}}

\newcommand{\tabincell}[2]{\begin{tabular}{@{}#1@{}}#2\end{tabular}}

\ifCLASSOPTIONcompsoc
\usepackage[nocompress]{cite}
\else
\usepackage{cite}
\fi
\ifCLASSINFOpdf
\else
\fi
\begin{document}
	\title{Transitional Learning: Exploring the Transition States of Degradation for Blind Super-resolution}
	
	\author{Yuanfei~Huang,
		Jie~Li,
		Yanting~Hu,
		Xinbo~Gao,~\IEEEmembership{Senior Member,~IEEE},\\
		and~Hua~Huang,~\IEEEmembership{Senior Member,~IEEE}
		
		\IEEEcompsocitemizethanks{
			\IEEEcompsocthanksitem Yuanfei Huang and Hua Huang are with the School of Artificial Intelligence, Beijing Normal University, Beijing, 100875, China. E-mail: \{yfhuang, huahuang\}@bnu.edu.cn.
			\IEEEcompsocthanksitem Jie Li is with the Video and Image Processing System Laboratory, School of Electronic Engineering, Xidian University, Xi'an 710071, China. E-mail: leejie@mail.xidian.edu.cn. 
			\IEEEcompsocthanksitem Yanting Hu is with the School of Medical Engineering and Technology, Xinjiang Medical University, Urumqi 830011, China. E-mail: yantinghu2012@gmail.com.
			\IEEEcompsocthanksitem Xinbo Gao is with the Chongqing Key Laboratory of Image Cognition, Chongqing University of Posts and Telecommunications, Chongqing 400065, China (E-mail: gaoxb@cqupt.edu.cn) and with the School of Electronic Engineering, Xidian University, Xi'an 710071, China (E-mail: xbgao@mail.xidian.edu.cn).
			\IEEEcompsocthanksitem (Corresponding author: Hua Huang.)
		}
	}

	
	\IEEEtitleabstractindextext{%
		\begin{abstract}
			Being extremely dependent on iterative estimation of the degradation prior or optimization of the model from scratch, the existing blind super-resolution (SR) methods are generally time-consuming and less effective, as the estimation of degradation proceeds from a blind initialization and lacks interpretable degradation priors. 
			To address it, this paper proposes a transitional learning method for blind SR using an end-to-end network without any additional iterations in inference, and explores an effective representation for unknown degradation.
			To begin with, we analyze and demonstrate the transitionality of degradations as interpretable prior information to indirectly infer the unknown degradation model, including the widely used additive and convolutive degradations.
			We then propose a novel Transitional Learning method for blind Super-Resolution (TLSR), by adaptively inferring a transitional transformation function to solve the unknown degradations without any iterative operations in inference. Specifically, the end-to-end TLSR network consists of a degree of transitionality (DoT) estimation network, a homogeneous feature extraction network, and a transitional learning module.
			Quantitative and qualitative evaluations on blind SR tasks demonstrate that the proposed TLSR achieves superior performances and costs fewer complexities against the state-of-the-art blind SR methods.
			The code is available at \href{https://github.com/YuanfeiHuang/TLSR}{github.com/YuanfeiHuang/TLSR}.
		\end{abstract}
		
		}
	
	\maketitle
	
	\IEEEdisplaynontitleabstractindextext
	\IEEEpeerreviewmaketitle
	\IEEEraisesectionheading{\section{Introduction}\label{sec:introduction}}
	\IEEEPARstart{S}{ingle} image super-resolution (SISR), aiming at reconstructing a high-resolution (HR) image from a degraded low-resolution (LR) image, is considered an ill-posed inverse problem.
	For decades, many studies have been proposed to solve this ill-posed problem, including interpolation based~\cite{Bicubic1981TASSP}, reconstruction based~\cite{MarquinaA2008JSC,DongW2011TIP} and example learning based~\cite{YangJ2010TIP,ZeydeR2010,HuY2016TIP,HuangY2018TIP} methods.
	
	Recently, with the development of the high-profile deep convolutional neural networks (CNNs), and to learn the co-occurrence priors or non-linear representations between HR-LR pairs from a large training datasets, deep learning based SISR methods have received considerable attention due to their excellent performance and real-time processing. As the first attempt, Dong~\etal\cite{DongC2014ECCV,DongC2016TPAMI} proposed SRCNN by stacking a shallow CNN to learn the non-linear mappings of LR-to-HR pairs, which outperforms most existing example-based SR methods and leads the trend of using end-to-end networks for SISR~\cite{ShiW2016CVPR,DongC2016ECCV,TaiY2017ICCV,TongT2017ICCV,LedigC2017CVPR,TaiY2017CVPR,LimB2017CVPRW,ZhangY2018ECCV,ZhangY2018CVPR,HuY2020TCSVT,HuangY2021TIP,ZhangY2021TPAMI}. 
	Nevertheless, as the degradation is generally synthesized under an assumed degradation prior (\eg, bicubic down-sampling), these deep learning based SR models were trained by optimizing on numerous synthetic HR-LR pairs with single degradation, and unfortunately show less effectiveness on other degradations.
	
	Thus, the integrated models with conditional information have been developed by training a single model on multiple/variational degradations, \eg, SRMD~\cite{ZhangK2018CVPR} and UDVD~\cite{XuY2020CVPR}. However, these integrated models need giving a degradation prior condition to guide the model to solve the corresponding degradation and show less effectiveness when the degradation prior is unknown.
	Then, blind super-resolution (BSR) has been raised to improve the generalization of the model on unknown degradations, and the corresponding methods can be divided into two branches: {\em integrated learning based methods} (\eg, DnCNN~\cite{ZhangK2017TIP}, VDN~\cite{YueZ2019NeurIPS},  IKC~\cite{GuJ2019CVPR} and DASR~\cite{WangL2021CVPR}) and {\em zero-shot learning based methods} (\eg, ZSSR~\cite{ShocherA2018CVPR}, DIP~\cite{UlyanovD2018CVPR}, KernelGAN~\cite{BellS2019NeurIPS} and MZSR~\cite{SohJW2020CVPR}).
	
	
	However, by iteratively estimating the prior information of degradations or optimizing the model from scratch, most of the existing integrated learning based and zero-shot learning based BSR methods are time-consuming and not controllable enough. Therefore, it is natural to raise two questions: (1) Is it possible to facilitate the inference/testing phase in a non-iterative way? (2) Is there a natural characteristic in degradations to improve the adaptability of models?
	
	(1) For the first issue, an end-to-end network is necessary to simultaneously estimate the degradation priors and reconstruct the degraded LR images. Thus, we need a sufficiently accurate degradation estimation without any additional iterative corrections.
	
	(2) For the second issue, we revisit and analyze the relations between various degradations, and find that most degradations are generally transition states of two given primary degradations. 
	Therefore, it is feasible to super-resolve a LR image with an unknown transitional degradation using the models trained only on the corresponding primary degradations. 
	
	\begin{figure}
		\centering
		\includegraphics[width=1\linewidth]{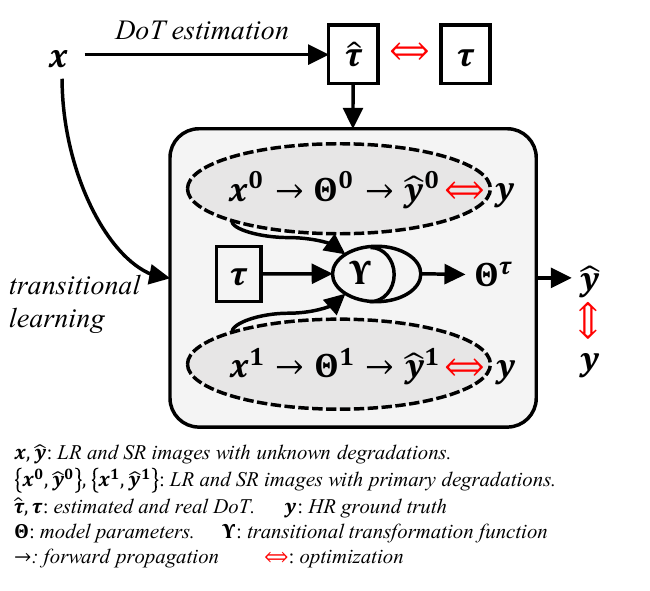}
		\caption{Overall flow of transitional learning for blind SR.}
		\label{fig:overall_flow}
	\end{figure}
	Inspired by these issues, as in Fig.~\ref{fig:overall_flow}, we explore the transitionality among degradations and propose a Transitional Learning method for blind Super-Resolution (TLSR). Our main contributions are as follows.
	
	$\bullet$ We analyze and demonstrate the transitionality of additive and convolutive degradation, which indicates that most unknown transition states of degradation can be represented by two or more primary degradations.
	
	$\bullet$ We propose a novel transitional learning method for blind SR, by adaptively rebuilding a transitional transformation function to solve the unknown degradations, which is faster and more accurate than the existing blind SR methods. For example, for $\times2$ upscaling on Set14 dataset with convolutive degradations, our TLSR gains PSNR of 0.69dB and 3.63dB, and is about $\times10$ and $\times2000$ times faster than IKC and ZSSR, respectively. 
	
	$\bullet$ We propose an assessment-aware module to estimate the degree of transitionality, which provides convincing prior information to guide the transitional learning module.
	
	$\bullet$ Theoretical and experimental demonstrations indicate that, the proposed TLSR achieves superior performance and efficiency against the state-of-the-art blind SR methods.
	
	\begin{figure*}
		\centering
		\includegraphics[width=1\linewidth]{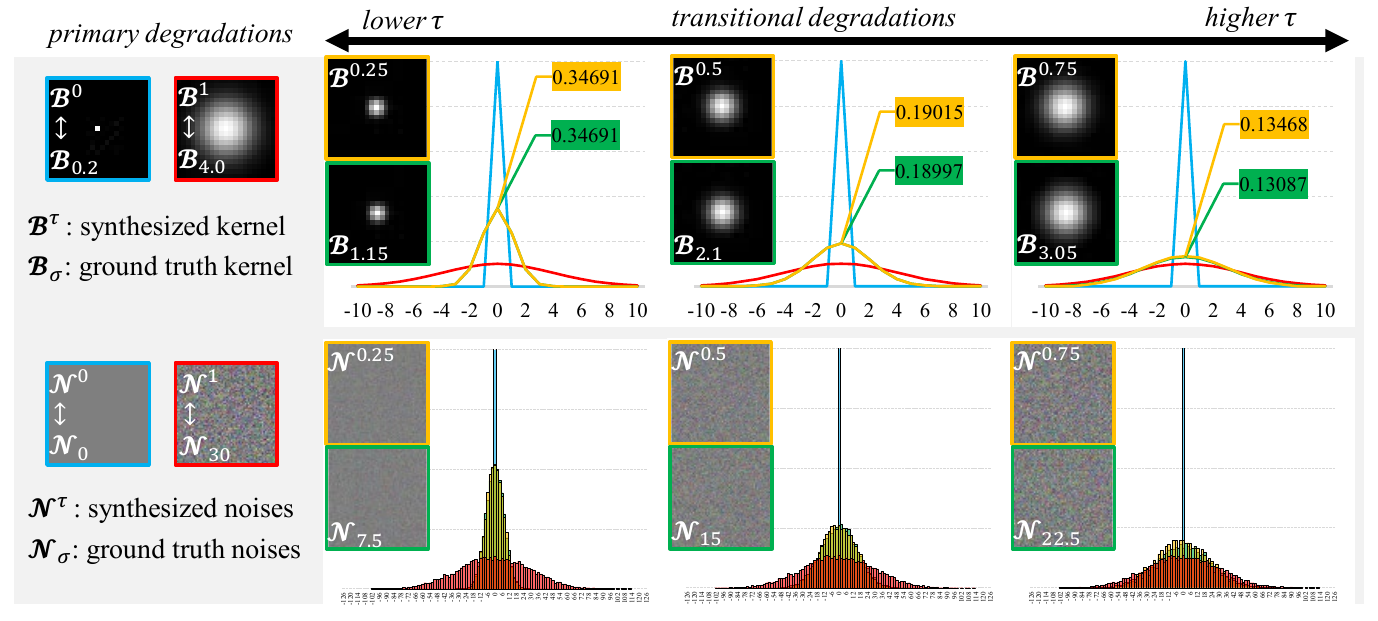}
		\caption{For both convolutive degradations and additive degradations, arbitrary degradation (blur kernel or noisy map) can be approximated by the transition state of two primary degradations.}
		\label{fig:transitionality_degradation}
	\end{figure*}
	\section{Related Work}
	\subsection{Non-Blind Super-Resolution}
	Non-blind super-resolution aims at reconstructing the degraded LR images using a single model trained with the given prior information of degradations and well-designed objective optimization functions, which is the same as SISR.
	Representatively, SRCNN~\cite{DongC2016TPAMI}, VDSR~\cite{KimJ2016CVPR_VDSR}, EDSR~\cite{LimB2017CVPRW}, SRGAN~\cite{LedigC2017CVPR} and ESRGAN~\cite{WangX2018ECCVW} were presented to learn the non-linear representation of LR-to-HR mappings using $\ell_2$/$\ell_1$ loss functions or adversarial learning, where the degraded LR images are generated by a single bicubic down-sampling degradation. 
	
	To improve the generalization of models on various degradations, SRMD~\cite{ZhangK2018CVPR} and UDVD~\cite{XuY2020CVPR} were proposed by integrating numerous degradations in a single integrated network, which is a conditional method for solving various degradations under the non-blind degradation prior. Besides, for arbitrary blur kernels, plug-and-play~\cite{ZhangK2019CVPR,WeiK2020ICML,ZhangK2021TPAMI}, and deep unfolding network~\cite{ZhangK2020CVPR} were introduced to correct the model using non-blind priors.
	\subsection{Blind Super-Resolution}
	However, the degradations are generally unknown in real-world applications. Since the non-blind models are trained on a priori of single or multiple specific degradations, they would show less effectiveness when the prior of degradation is not given. Thus, blind super-resolution has been raised to solve this issue for a better generalization on real-world scenes, and could be divided into two branches according to the manner of optimization as follows.
	\subsubsection{Integrated learning based methods}
	For unknown degradations, based on a non-blind integrated model $\boldsymbol{\mathcal{F}}_\Theta$, which is trained with variational degradations and the corresponding given conditional information of degradation prior $\mathcal{D}$, the pipeline of integrated learning based BSR methods (IL-BSR) is to predict $\mathcal{D}$ by an estimator $\boldsymbol{\mathcal{F}}_\mathcal{D}$ firstly and then utilize it as conditional guidance of $\boldsymbol{\mathcal{F}}_\Theta$ to estimate a desired result as
	\begin{equation}
		\left\{\begin{array}{lr}
			\mathcal{D}_k=\boldsymbol{\mathcal{F}}_\mathcal{D}(\boldsymbol{y}_{k-1}) &  \\
			\boldsymbol{y}_k=\boldsymbol{\mathcal{F}}_\Theta(\boldsymbol{y}_{k-1}, \mathcal{D}_k ) &  
		\end{array}\right.
	\end{equation}
	where $\boldsymbol{y}_{k}$ denotes the current super-resolved result, and $\boldsymbol{y}_{0}$ indicates the LR input $\boldsymbol{x}$.
	
	Representatively, IKC~\cite{GuJ2019CVPR} was proposed to iteratively predict and correct the conditional information of degradation using the LR inputs or super-resolved results, and apply the estimated conditional information into an integrated non-blind model. Furthermore, DASR~\cite{WangL2021CVPR} utilized the contrastive learning~\cite{HeK2020CVPR} to estimate a discriminative degradation representation as conditional information. Moreover, toward blind image denoising, VDN~\cite{YueZ2019NeurIPS} was proposed by exploiting a deep variational inference network for noise modeling and removal. However, these IL-BSR methods are not controllable enough, as they largely depend on the estimated $\mathcal{D}$ which is high-dimension vectors and with high freedom for representation. 
	
	\subsubsection{Zero-shot learning based methods}
	Except for modeling variational degradations by a single model, the zero-shot learning based BSR methods (ZSL-BSR) have been raised to generate a specific model using a single image for an unknown degradation. Without the ground truth, ZSL-BSR iteratively optimize the image-specific model $\boldsymbol{\mathcal{F}}_{\Theta_{k}}$ under the self-supervised guidances, as
	\begin{equation}
		\left\{\begin{array}{lr}
			\Theta_k = \arg\min\limits_{\Theta}\boldsymbol{\mathcal{L}}(\boldsymbol{y_{k-1}},\boldsymbol{x}, \boldsymbol{\mathcal{F}_{\Theta}})  &  \\
			\boldsymbol{y}_k=\boldsymbol{\mathcal{F}}_{\Theta_{k}}(\boldsymbol{y}_{k-1}) &  
		\end{array}\right.
	\end{equation}
	where $\boldsymbol{\mathcal{L}}$ represents a predefined objective function.

	Representatively, without any external dataset, ZSSR~\cite{ShocherA2018CVPR} was exploited to solve image-specific blind SR using internal patches of a single image. Similarly, without explicit prior information of degradation, DIP~\cite{UlyanovD2018CVPR} utilized a deep network to learn the image priors for image restoration; Noise2Noise~\cite{LehtinenJ2018ICML} applied statistical reasoning by corrupting the corrupted observations for image denoising; KernelGAN~\cite{BellS2019NeurIPS} was explored to estimate the blur kernels and use the estimated kernels to generate degraded images for further training or inference. Besides, meta-learning~\cite{FinnC2017ICML} was introduced to facilitate model adaptation, \eg, MZSR~\cite{SohJW2020CVPR} and MLSR~\cite{ParkS2020ECCV}. However, these methods are time-consuming as they require iterative optimization during inference.
	
	\subsection{Preliminary Definition of Transition State}

	In this paper, we introduce the conception of transition state as an intermediate variable from a primary state to other states, particularly in a continuous system, it can be formulated by finite primary states. 	
	Particularly, similar conceptions of transition state have important significance in many fields of science and engineering, as the transition states are mainly used to represent some intermediate processes or states of continuous systems. For example, in the color spectrum of computer graphics, a continuous colormap~\cite{Nardini2019TVCG,BujackR2017TVCG} indicates a series of transitional colors generated by mixing the primary colors, \eg, red, green, and blue. In latent code representation of variational auto-encoder~\cite{KingmaD2014ICLR_VAE}, it is feasible to explore a transitional manifold for implicit image generation and representation~\cite{MeschederL2017ICML}.
	
	Besides, transition states also exist in other fields, such as denoising diffusion probabilistic model~\cite{HoJ2020NeurIPS} in image generation, intermediate time evolution~\cite{Gossom2016PhysReports} in the theorem of Schr{\"o}dinger picture~\cite{Dirac1964Nature} and phyletic gradualism~\cite{Rhodes1983Nature} in evolutionary biology.
	
	\begin{figure*}
		\centering
		\includegraphics[width=1\linewidth]{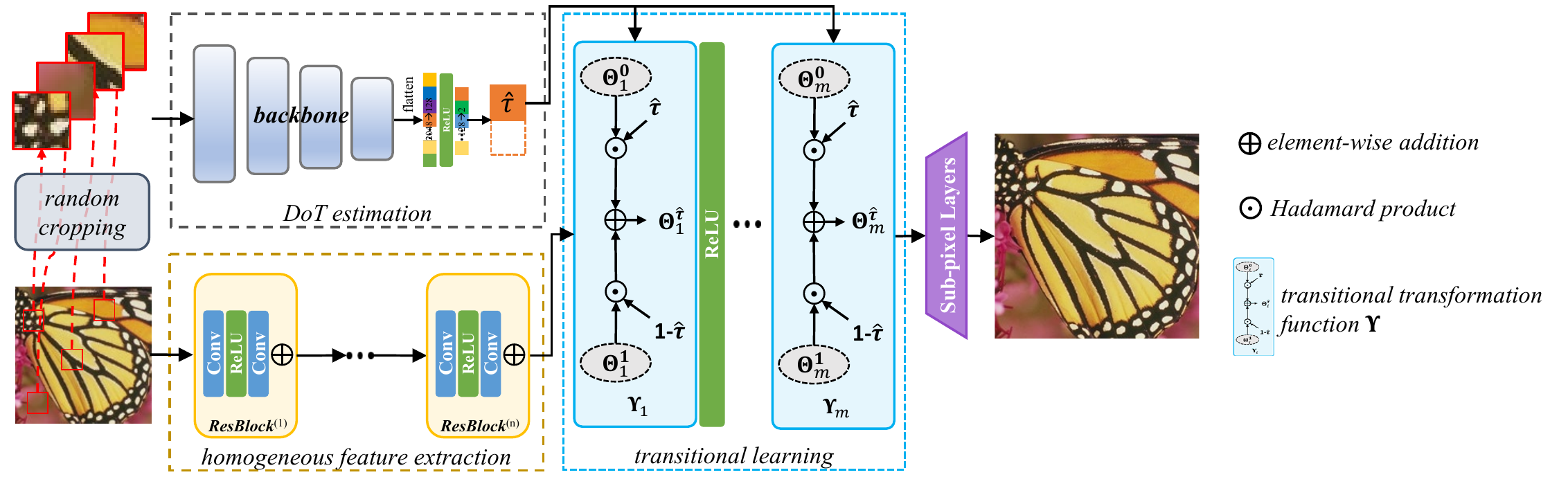}
		\caption{Architecture of the proposed TLSR, which consists of three modules: (1) homogeneous feature extraction network, which is parameter-shared for arbitrary transitional degradations; (2) DoT estimation network for estimating $\hat \tau$ from the randomly cropped local patches; (3) transitional learning module stacks a group of transitional transformation function $\boldsymbol{\Upsilon}$ to rebuild an adaptive network.}
		\label{fig:architecture}
	\end{figure*}
	\section{Proposed Method}
	Inspired by {\em transition state}, if the given degradation is transitional, arbitrary unknown variants would be represented by two or more primary states. So, to begin with, we analyze the transitionality of the widely used additive and convolutive degradations. Then we describe the methodologies of transitional learning and its derived implementation on blind super-resolution.
	
	\subsection{Transitionality of Degradations}\label{sec:transitionality of degradations}
	In classic SR problem, there generally exists two types of degradation (convolutive degradation $\mathcal{B}$ for isotropic Gaussian blur and additive degradation $\mathcal{N}$ for AWGN) in the LR corrupted images, which is formulated as
	\begin{equation}
		\boldsymbol{x}= (\boldsymbol{y}\otimes\mathcal{B})\downarrow_{\times s}+\mathcal{N}
		\label{eq:degradation model}
	\end{equation}
	where $\otimes$ represents the convolution operator, $\downarrow_{\times s}$ indicates $\times s$ downsampling (\eg, bicubic downsampling), $\boldsymbol{x}$ and $\boldsymbol{y}$ denote LR image and its HR counterpart respectively.
	
	Aiming at solving the SR task on this complex degradation, most existing non-blind and blind SR methods attempt to optimize a well-designed specific or integrated model $\boldsymbol{\mathcal{F}}(\boldsymbol{x};\Theta, \uparrow_{\times s}):\boldsymbol{x}\mapsto\boldsymbol{y}$. For a specific degradation in problem space, the accurate degradation priors make the optimization of models more effective. However, these models would be less effective if a degradation is unseen in the problem space. Therefore, we expect to find an effective representation for arbitrary degradation in or beyond the problem space, including additive degradations and convolutive degradations.
	
	\subsubsection{Additive degradation $\mathcal{N}$}
	Since the additive degradations are independent and linearly additive, given the degree of transitionality (DoT) $\tau$, the transition states $\{\boldsymbol{x}^{\tau}\}$ are then formulated as
	\begin{equation}
		\begin{aligned}
		\boldsymbol{x}^{\tau}&=\boldsymbol{x} + \mathcal{N}^\tau\\
		&=\tau(\boldsymbol{x}+\mathcal{N}^0)+(1-\tau)(\boldsymbol{x}+\mathcal{N}^1)
		\end{aligned}
		\label{eq:transitionalfunction_N}
	\end{equation}
	where, $\mathcal{N}^\tau=\tau\mathcal{N}^0 + (1-\tau)\mathcal{N}^1$ represents that the transitional additive degradations can be represented by two primary degradations $\mathcal{N}^0$ and $\mathcal{N}^1$. Particularly, when the original state is noise-free ($\mathcal{N}^0=0$), the estimation of transition state is simplified into a linear polynomial function as $\mathcal{N}^\tau=(1-\tau)\mathcal{N}^1$.
	
	\subsubsection{Convolutive degradation $\mathcal{B}$} \label{sec:transitionality_conv}
	Unlike the additive degradations, the convolutive degradations are generated by non-linear functions (\eg, Gaussian kernels) and not linearly additive. Nevertheless, inspired by the Gaussian mixture model (GMM)~\cite{HuangY2018TIP}, we can infer any convolutive Gaussian distribution as a mixture of multiple Gaussian distributions as
	\begin{equation}
		\mathcal{B}=\sum_{i=0}^{K-1}\tau_i\mathcal{B}^i\sim\sum_{i=1}^{K-1}\tau_i\mathbb{N}(\mu_i, \sigma^2_i)
	\end{equation}
	particularly, when $K=2$, assume that the primary convolutive degradations are independent and $\mathcal{B}^0\sim \mathbb{N}(0, \sigma^2_0)$, $\mathcal{B}^1\sim \mathbb{N}(0, \sigma^2_1)$. Specifically in {\em Proposition}~\ref{proposition:1}, the transition state of convolutive degradation can be formulated as an algebra combination of the primary convolutive degradations, which demonstrates the transitionality of convolutive degradation.
	
	\begin{proposition}
		Given two Gaussian blur kernels $\mathcal{B}^0\sim\mathbb{N}(0, \sigma^2_0)$, $\mathcal{B}^1\sim\mathbb{N}(0, \sigma^2_1)$ ($\sigma_1>\sigma_0$), and degree of transitionality $\tau$, then in element-wise $(i,j)$, $\mathcal{B}^\tau\sim\mathbb{N}(0, \sigma^2_\tau)$ can be represented as a transition state between $\mathcal{B}^0$ and $\mathcal{B}^1$:
		\begin{equation}
			\mathcal{B}^\tau(i,j) \propto (\mathcal{B}^0(i,j))^{\frac{\sigma^2_0}{2\sigma_\tau^2}}(\mathcal{B}^1(i,j))^{\frac{\sigma^2_1}{2\sigma_\tau^2}}
		\end{equation}
		where $\sigma_\tau=(1-\tau)\sigma_0+\tau\sigma_1$.
		See proofs in Appendix~\ref{appendix:proof}.
		\label{proposition:1}
	\end{proposition}
	
	Specifically, when a primary degradation $\mathcal{B}^0$ is blur-free ($\sigma_0=0^+$), the transition state $\{\boldsymbol{x}^{\tau}\}$ are formulated as
	\begin{equation}
		\boldsymbol{x}^{\tau}\propto\boldsymbol{x}\otimes (\mathcal{B}^1)^{\frac{1}{2\tau^2}}
	\end{equation}
	namely, the estimation of transition states is simplified as a variant of $\mathcal{B}^1$.
	
	Furthermore, if $\sigma_0\neq0$, the transitional degradation is a non-linear combination of $\mathcal{B}^0$ and $\mathcal{B}^1$, and the transition state is transformed into
	\begin{equation}
		\boldsymbol{y}^{\tau}\propto\boldsymbol{y}\otimes ((\mathcal{B}^0)^\frac{\sigma^2_0}{2\sigma_\tau^2})(\mathcal{B}^1)^\frac{\sigma^2_1}{2\sigma_\tau^2}))
		\label{eq:transition_state_conv}
	\end{equation}
	
	Therefore, as Fig.~\ref{fig:transitionality_degradation} shows, both additive and convolutive degradations are transitional. Thus, it is feasible to represent an unknown degradation by the transition state between two or more primary degradations.
	
	\subsection{Transitional Learning}
	To solve the SR problem of a LR image $\boldsymbol{x}$, it allows learning the energy function by {\em Maximum A Posteriori} (MAP) as
	\begin{equation}
		\begin{aligned}
		\boldsymbol{\hat y} &= \arg\max_{\boldsymbol{y}} \log P(\boldsymbol{y}|\boldsymbol{x})\\
		&= \arg\max_{\boldsymbol{y}} \log P(\boldsymbol{x}|\boldsymbol{y}) + \log P(\boldsymbol{y})
		\end{aligned}
	\end{equation}
	particularly, $P(\boldsymbol{x}|\boldsymbol{y})$ indicates the degradation prior and is commonly unknown in blind tasks. 
	
	To solve this issue, if we assume that the combination of primary states $\{\boldsymbol{x}^i\}^{K-1}_{i=0}$ is a complete set for representing an unknown state $\boldsymbol{x}$, then as $\{\boldsymbol{x}^i\}^{K-1}_{i=0}$ are conditionally independent, we would formulate the transitional degradation as
	\begin{equation}
			P(\boldsymbol{x}|\boldsymbol{y}) = P(\boldsymbol{x}^0,\boldsymbol{x}^1,...,\boldsymbol{x}^{K-1}|\boldsymbol{y})
			=\prod_{i=0}^{K-1} P(\boldsymbol{x}^i|\boldsymbol{y})
	\end{equation}
	and
	\begin{equation}
	\hat{\boldsymbol{y}} = \arg\max_{\boldsymbol{y}} \sum_{i=0}^{K-1}\log P(\boldsymbol{x}^i|\boldsymbol{y}) + \log P(\boldsymbol{y})
	\end{equation}
	so we can infer the optimal $\hat{\boldsymbol{y}}$ under the prior of $K$ given primary degradation priors $\{P(\boldsymbol{x}^i|\boldsymbol{y})\}^{K-1}_{i=0}$.
	
	Therefore, when the degradation fulfills transitionality as illustrated in Section~\ref{sec:transitionality of degradations}, although without precise information of the unknown degradation $\mathcal{B}$ and $\mathcal{N}$, it is feasible to infer a desired super-resolved result by finite independent primary degradations $\{\mathcal{B}^i\}^{K-1}_{i=0}$, $\{\mathcal{N}^i\}^{K-1}_{i=0}$ and the corresponding DoT $\{\tau_i\}^{K-1}_{i=0}$. 
	
	Ultimately, under this principle, we introduce a {\em transitional learning} method to infer an unknown transitional degradation as Fig.~\ref{fig:overall_flow} and Fig.~\ref{fig:architecture} show, which attempts to infer the posteriori on a transition state $\boldsymbol{x}$ by finite posteriori on primary states $\{\boldsymbol{x}\}^{K-1}_{i=0}$, and formulates the optimal inference model $\boldsymbol{\mathcal{F}}(\boldsymbol{x};\Theta, \uparrow_{\times s}):\boldsymbol{x} \mapsto\boldsymbol{y}$ with 
	\begin{equation}
		\Theta=\boldsymbol{\Upsilon}(\{\Theta^i\}^{K-1}_{i=0}, \{\tau_i\}^{K-1}_{i=0})
		\label{eq:transformation}
	\end{equation}
	where $\boldsymbol{\Upsilon}$ denotes a transitional transformation function for rebuilding a new model. 
	
	Particularly, from the demonstration on transitionality, we need only two primary degradations to represent the transitional degradations $\mathcal{B}$ and $\mathcal{N}$. Thus, we consider a specific case (when $K=2$) as 
	\begin{equation}
		\Theta^\tau=\boldsymbol{\Upsilon}(\Theta^0,, \Theta^1, \tau)
		\label{eq:transformation_N}
	\end{equation}
	and next apply it into blind SR with additive and convolutive degradations in detail. 
	
	\begin{figure}[!t]
		\centering
		\includegraphics[width=0.9\linewidth]{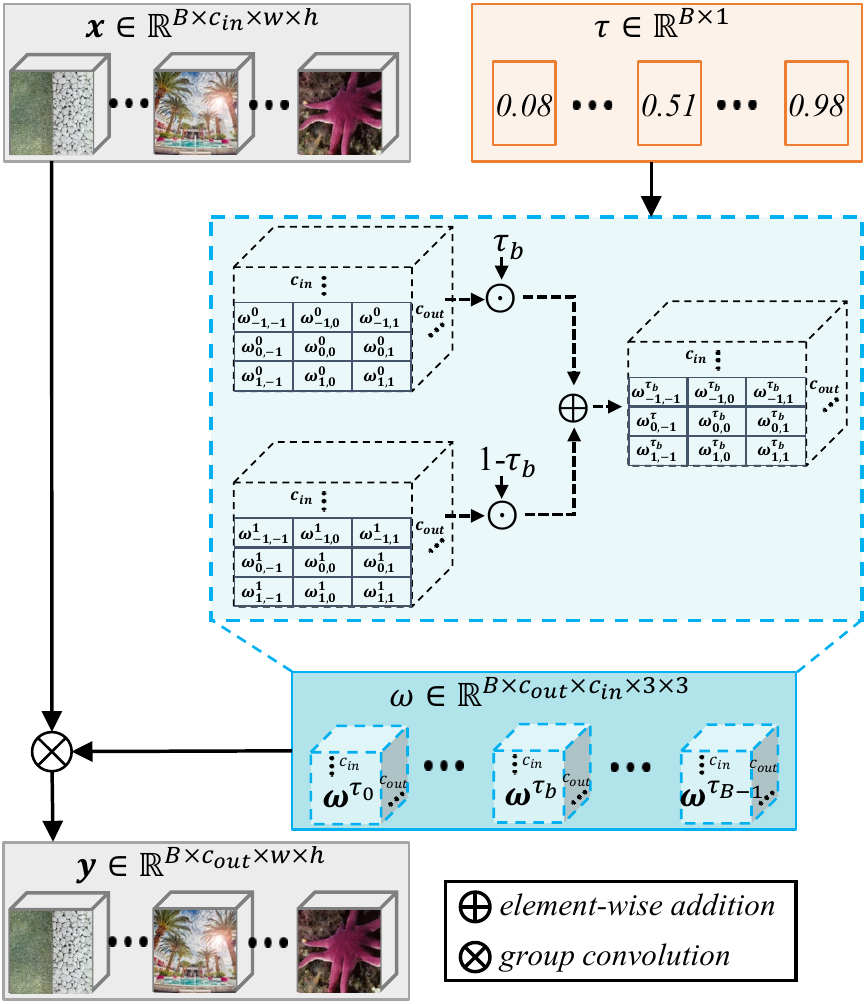}
		\caption{Network interpolation: a feasible implementation of the transitional transformation function $\Upsilon$. Unlike~\cite{WangX2019CVPR}, since each sample in a batch has different DoT $\tau_b$, we apply each specific $\tau_b$ to rebuild the corresponding adaptive network and use the group convolutions for acceleration.}
		\label{fig:transitional_modal_learning}
	\end{figure}
	\subsubsection{Additive degradation $\mathcal{N}$}
	According to Eq.(\ref{eq:transitionalfunction_N}) and considering additive degradations only, the transition state $\boldsymbol{x}^{\tau}$ could be obtained through an explicit interpolation of two prior LR images $\boldsymbol{x}^0$ and $\boldsymbol{x}^1$ as
	\begin{equation}
		\boldsymbol{x}^{\tau}=\boldsymbol{y}\downarrow_{\times s} + \mathcal{N}^{\tau}
		=\tau\boldsymbol{x}^0 + (1-\tau)\boldsymbol{x}^1
		\label{eq: additive_degrad}
	\end{equation}
	where $\boldsymbol{x}^0=\boldsymbol{y}\downarrow_{\times s}+\mathcal{N}^0$ and $\boldsymbol{x}^1=\boldsymbol{y}\downarrow_{\times s}+\mathcal{N}^1$ represent the states by applying primary degradations respectively.
	
	However, since the deep-learning-based SR models consist of non-linear modules (Conv+ReLU), the super-resolved results $\boldsymbol{y}$ can not be estimated by explicitly linear interpolation of $\boldsymbol{y}^0=\boldsymbol{\mathcal{F}}(\boldsymbol{x}^0;\Theta^0)$ and $\boldsymbol{y}^1=\boldsymbol{\mathcal{F}}(\boldsymbol{x}^1;\Theta^1)$.
	Therefore, in conventional deep-learning-based SR methods, a well-designed CNN model is generally applied to super-resolve $\boldsymbol{x}^{\tau}$ as $\boldsymbol{\mathcal{F}}(\boldsymbol{x}^\tau;\Theta^\tau):\boldsymbol{x}^\tau\rightarrow\boldsymbol{y}$, but it is inefficient to train numerous models for all $\tau\sim\mathbb{U}(0,1)$ in practical applications.
	
	In particular, inspired by the network interpolation~\cite{WangX2018ECCVW,WangX2019CVPR} which uses the linear interpolation of convolutional layers to rebuild a new network, we revisit it and exploit its feasibility to the non-linear representation of transitional state. Thus, given two convolutional layers $\boldsymbol{\mathcal{F}}^0(\cdot;\Theta^0)$ and $\boldsymbol{\mathcal{F}}^1(\cdot;\Theta^1)$, we introduce a transitional learning module to derive a transitional model $\boldsymbol{\mathcal{F}}^{\tau}(\cdot;\Theta^{\tau})$ using a transitional transformation function $\boldsymbol{\Upsilon}$. In short, by utilizing the network interpolation in Fig.~\ref{fig:transitional_modal_learning}, we have $\Theta^{\tau}={\tau}\Theta^0+(1-{\tau})\Theta^1$. 
	
	Specifically in convolutional layers (with no bias), assume that the transitional LR image $\boldsymbol{x}^{\tau}$ can be accurately estimated by the interpolation of $\boldsymbol{x}^0$ and $\boldsymbol{x}^1$. We then infer the transitional results by feeding $\boldsymbol{x}^{\tau}$ into such transitional network as
	\begin{equation}
		\begin{aligned}
			\boldsymbol{\hat y}_{j}(p_o)&=\sum\limits_{i=1}^{c_{in}}{\sum\limits_{p\in \mathcal{R}}{\boldsymbol{x}^{\tau}_{i}(p_o+p)\cdot \boldsymbol{w}^{\tau}_{j,i}(p)}}\\
			&=\sum\limits_{i=1}^{c_{in}}{\sum\limits_{p\in \mathcal{R}}{({\tau}\boldsymbol{x}^0_{i}(p_o+p)+(1-{\tau})\boldsymbol{x}^1_{i}(p_o+p))}} \cdot \\
			&\ \ \ \ \ \ \ \ \ \ \ \ \ \ \ \ \ \ ({\tau}\boldsymbol{w}^0_{j,i}(p)+(1-{\tau})\boldsymbol{w}^1_{j,i}(p))\\
			&=\sum\limits_{i=1}^{c_{in}}{\sum\limits_{p\in \mathcal{R}}{{\tau}^2\boldsymbol{x}^0_{i}(p_o+p)\cdot\boldsymbol{w}^0_{j,i}(p)}}+\\
			&\ \ \ \ \ \ \ \ \ \ \ \ \ \ \ \ \ \ {\tau}(1-{\tau})\boldsymbol{x}^1_{i}(p_o+p)\cdot\boldsymbol{w}^0_{j,i}(p)+\\
			&\ \ \ \ \ \ \ \ \ \ \ \ \ \ \ \ \ \ {\tau}(1-{\tau})\boldsymbol{x}^0_{i}(p_o+p)\cdot\boldsymbol{w}^1_{j,i}(p)+\\
			&\ \ \ \ \ \ \ \ \ \ \ \ \ \ \ \ \ \ (1-{\tau})^2\boldsymbol{x}^1_{i}(p_o+p)\cdot\boldsymbol{w}^1_{j,i}(p)
		\end{aligned}
	\end{equation}
	where $\boldsymbol{w}$ represents the kernel weights of convolution, moreover, $c_{in}$ is the number of channels in input images (or features), $p_o$ is the initial position in images/features, and $p$ is the position of convolution in receptive filed $\mathcal{R}$ respectively.
	
	More generally, from Eq.(\ref{eq: additive_degrad}), the transition state $\boldsymbol{x}^{\tau}$ can be accurately estimated by the interpolation of $\boldsymbol{x}^0$ and $\boldsymbol{x}^1$, we then infer the results by feeding $\boldsymbol{x}^{\tau}$ into such transitional model as
	\begin{equation}
		\begin{aligned}
			\boldsymbol{\hat y}&=\boldsymbol{\mathcal{F}}^{\tau}(\boldsymbol{x}^{\tau};{\Theta}^{\tau})\\ &={\tau}^2\boldsymbol{\mathcal{F}}^0(\boldsymbol{x}^0;{\Theta}^0)+{\tau}(1-{\tau})\boldsymbol{\mathcal{F}}^1(\boldsymbol{x}^0;{\Theta}^1)+\\
			&\ \ \ \ \ {\tau}(1-{\tau})\boldsymbol{\mathcal{F}}^0(\boldsymbol{x}^1;{\Theta}^0)+(1-{\tau})^2\boldsymbol{\mathcal{F}}^1(\boldsymbol{x}^1;{\Theta}^1)
		\end{aligned}
	\end{equation}
	this transitional model shows a substantial capacity of feature representation with more than 4 SISR effects.
	
	\subsubsection{Convolutive degradation $\mathcal{B}$}
	As in Eq. (\ref{eq:transition_state_conv}), the transition state of convolutive degradations $\mathcal{B}^\tau$ can be explicitly represented by two primary degradations $\mathcal{B}^0$ and $\mathcal{B}^1$.
	Similar to the transitional transformation function of additive degradations in Eq.(\ref{eq:transformation_N}), we expect to explore a transitional transformation function to approximate the transition state in convolutive degradations. 
	
	However, in Eq.(\ref{eq:transition_state_conv}), unlike the additive degradations, a transition state of convolutive degradation is a non-linear transformation of the primary states.
	Intuitively, as in Fig.~\ref{fig:architecture}, we design a deep network with non-linear activation to approximately fit this transitional transformation function as
	\begin{equation}
		\boldsymbol{y}=\boldsymbol{\mathcal{F}}_{m}^{\tau}(... \mathcal{S}(\boldsymbol{\mathcal{F}}_1^{\tau}(\boldsymbol{x}^{\tau};\Theta_1^{\tau}));\Theta_{m}^{\tau})\\
	\end{equation}
	where $\mathcal{S}$ represents a non-linear activation function, \eg, ReLU. For generalization, we use this non-linear network to fit the transitional transformation function for both additive and convolutive degradations.
	
	Thus, for a mini-batch of size $B$, the objective function of our TLSR is formulated as
	\begin{equation}
		\boldsymbol{\mathcal{L}}_\text{TLSR}=\frac{1}{B}\sum_{b=1}^{B}\left \| {\boldsymbol{\mathcal{F}}}_\text{TLSR}(\boldsymbol{x}_b;\Theta_\text{TLSR}^{\tau_b})-\boldsymbol{y}_b \right \|_1
	\end{equation}
	where $\boldsymbol{y}_b$ denotes the HR ground truth.
	
	Specifically, as Fig.~\ref{fig:architecture} shows, the LR images are firstly fed into a homogeneous feature extraction network and parameter-shared for degradations with any DoTs. Then, the homogeneous features are fed into the transitional learning module, which stacks $m$ transitional transformation functions to infer an adaptive model for the transitional degradation.
	Furthermore, since the tenability of this conclusion is under the assumption that the transition state can be well represented using the real DoT, it is crucial to estimate DoT $\tau$, and make the transitional learning module more adaptive to blind SR task. 
	
	\subsection{DoT Estimation}\label{sec:DPNet}
	As above, the transitional model can be formulated as the interpolation of finite primary models. Nevertheless, since the real DoT $\tau$ is generally unknown in blind SR tasks, estimating the desired transitional model is difficult. Thus in this section, we mainly describe an effective way to estimate an appropriate DoT $\hat\tau$, the architecture is shown in Fig.~\ref{fig:architecture} and named DoTNet.
	
	\subsubsection{Random cropping}
	Since DoTNet is designed as a score regression network with fully connected layers (FC) in tail, fixed-size image patches need to be fed into the DoTNet. Therefore, we first introduce a random cropping module to obtain a group of fixed-sized patches from the input LR image. The reasons for patch-cropping are that: (1) in practice, the image sizes are generally unfixed and (2) either convolutive or additive degradation acts on a local receptive field, then each local patch in a single image commonly have the same representation for degradation.
	
	Specifically, taking convolutive degradations (\eg, Gaussian blur effects) into consideration, Gaussian blur kernel of size $w$ is sliding on each pixel of patch. Since the convolutive degradation has a sliding receptive field of size $w$, to capture the pattern of degradation effects, we need fixed-size image patches with $\min(width, height)\ge w$. Besides, as the considered AWGN is independent of the image and acts on isolated pixels, we only need $\min(width, heright)\ge 1$ to capture the pattern of additive degradations.
	
	\subsubsection{DoTNet}
	Given groups of LR image patches with variational degradations $\{\{\boldsymbol{x}_{b,t}\}_{t=1}^T\}_{b=1}^B$, the output DoT is obtained by capturing the activation in the last layer as $\hat\tau=\boldsymbol{\mathcal{F}}_\text{DoT}(\boldsymbol{x};\Theta_\text{DoT})$.
	We then train the parameters $\Theta_\text{DoT}$ by optimizing the objective function:
	\begin{equation}
		\boldsymbol{\mathcal{L}}_\text{DoT}=\frac{1}{B}\frac{1}{T}\sum_{b=1}^{B}\sum_{t=1}^{T}\left \| \boldsymbol{\mathcal{F}}_\text{DoT}(\boldsymbol{x}_{b,t};\Theta_\text{DoT})-\tau_b \right \|_1
	\end{equation}
	where $\tau_b$ indicates the real DoT of all the patches $\{\boldsymbol{x}_{b,t}\}_{t=1}^T$ from $b$-th LR image, and is preset by inferring the transition state mathematically as Section~\ref{sec:transitionality of degradations}.
	
	Specifically, we design the DoT estimation module as an assessment-aware score regression network~\cite{ZhuH2020CVPR} as in Fig.~\ref{fig:architecture}, where the feature extraction part (backbone) stacks 4 residual bottleneck blocks~\cite{HeK2016CVPR} with pooling operations. Moreover, in the decision part, we stack two fully connected layers and a Sigmoid function with a one-dimension output to estimate the DoT $\hat\tau$.
%
	
	\begin{figure*}
		\centering
		\includegraphics[width=0.8\linewidth]{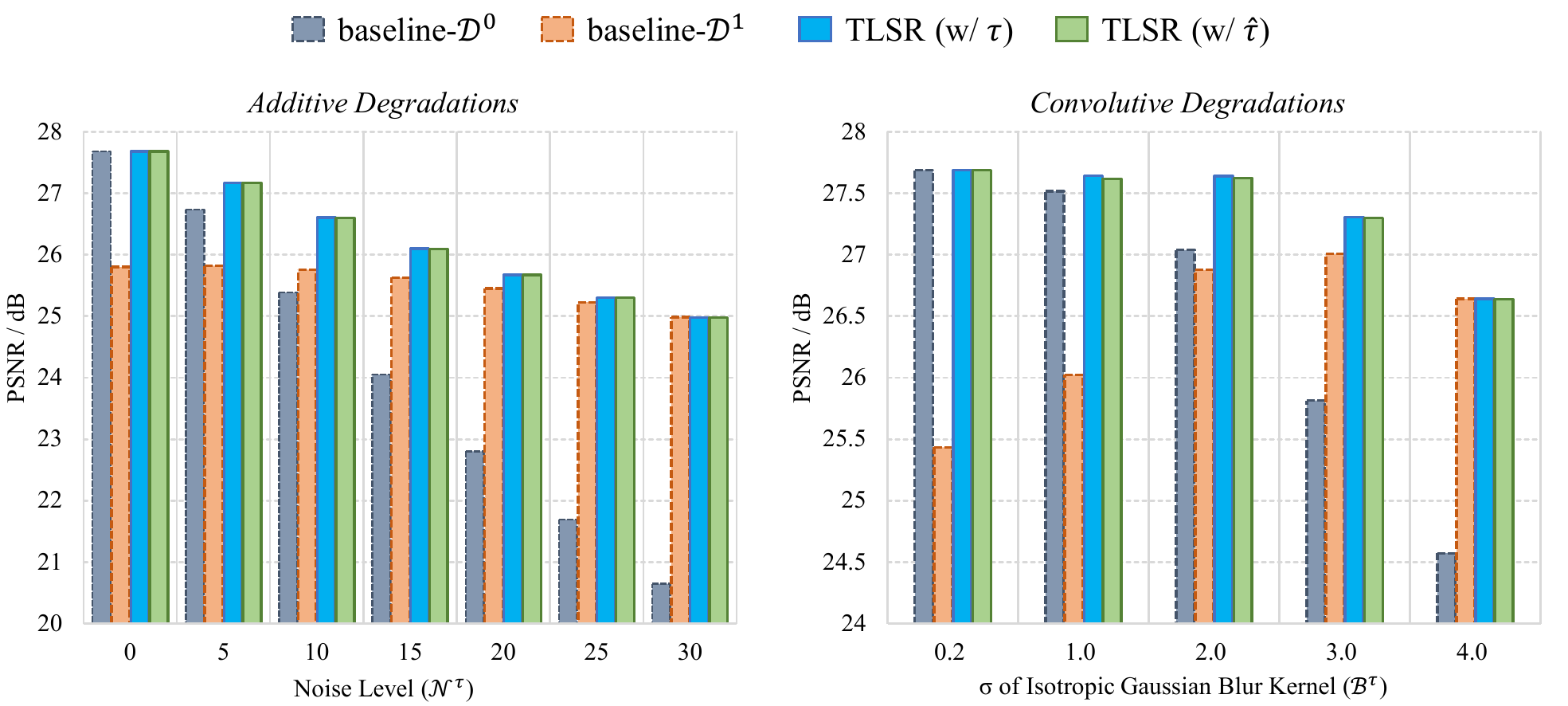}
		\caption{Ablation studies of the transitional learning, the experiments are conducted on BSD100 dataset for $\times4$ upscaling.}
		\label{fig:ablation_study}
	\end{figure*}
	
	\subsection{Discussions on Transitional Learning}
	\subsubsection{Difference to integrated learning}
	Integrated learning based SR methods aim at training a single integrated model with multiple prior degradations~\cite{ZhangK2018CVPR}, and improving the generalization of a model with the capability of processing all the prior degradation effectively. For blind SR, the IL-BSR methods~\cite{GuJ2019CVPR,YueZ2019NeurIPS} additionally exploit a degradation prediction model to correct the unseen degradation into a prior degradation for SR inference. However, as only need to train two primary degradations and exploit the transitionality of degradations to infer the unknown transition states, the proposed transitional learning is more flexible than the IL-BSR methods which need to cover all the prior degradations.
	In fact, for these IL-BSR methods, a higher dimension of degradation representation indicates higher degree of freedom, and it is challenging to learn an accurate solution. 
	
	Furthermore, these IL-BSR methods aim to explore a large solution space to cover the problem space of multiple degradations, but aimlessly enlarging the solution space is lack of flexibility and inefficient. However, our TLSR attempts to explore an adaptive and new solution space by an interpretable nature of the transitional degradations.
	\subsubsection{Difference to zero-shot learning}
	Zero-shot learning based SR methods depend on the internal samples of a single image, and tend to train an image-specific model with a specific prior degradation. For blind SR, the ZSL-BSR methods~\cite{BellS2019NeurIPS,ParkS2020ECCV,ShocherA2018CVPR,UlyanovD2018CVPR} aim at iteratively optimizing the model with a specific initial degradation to adapt the internal samples with unseen degradations, so they suffer from redundant iterations and are highly dependent on the initial degradation. However, the proposed transitional learning is an end-to-end network without additional iterations and only needs the type of degradation, which is more efficient than the ZSL-BSR.
	
	\section{Experiments}
	\subsection{Datasets and Evaluation}
	In the training phase, high-quality images with 2K resolution from DIV2K and Flickr2K~\cite{TimofteR2017CVPRW} datasets are considered as HR ground truth, and are then downsampled with specific degradations (\eg, bicubic, isotropic Gaussian blur, AWGN) to generate the corresponding LR images. 
	In the testing phase, we conduct the experiments on several benchmark datasets for evaluation, including Set5~\cite{BevilacC2012BMVC}, Set14~\cite{ZeydeR2010}, BSD100~\cite{ArbelaezP2011TPAMI}, Urban100~\cite{HuangJB2015CVPR} with general and structural scenes.
	
	To evaluate SR performance, we apply two common full-reference image quality assessment criteria to evaluate the discrepancies of super-resolved results and HR ground truth, \ie, peak signal-to-noise ratio (PSNR) and structural similarity (SSIM). Following the convention of super-resolution, the luminance channel is selected for full-reference image quality assessment because the intensity of the image is more sensitive to human vision than the chroma. Furthermore, following~\cite{ZhangK2019AIM}, the inference time is calculated via \texttt{torch.cuda.Event} to evaluate the computational complexities on GPU.
	
	\begin{figure*}
		\centering
		\subfloat[Distribution of the estimated DoTs on BSD100 with convolutive degradations $\mathcal{B}^\tau$.]{
			\includegraphics[width=0.64\linewidth]{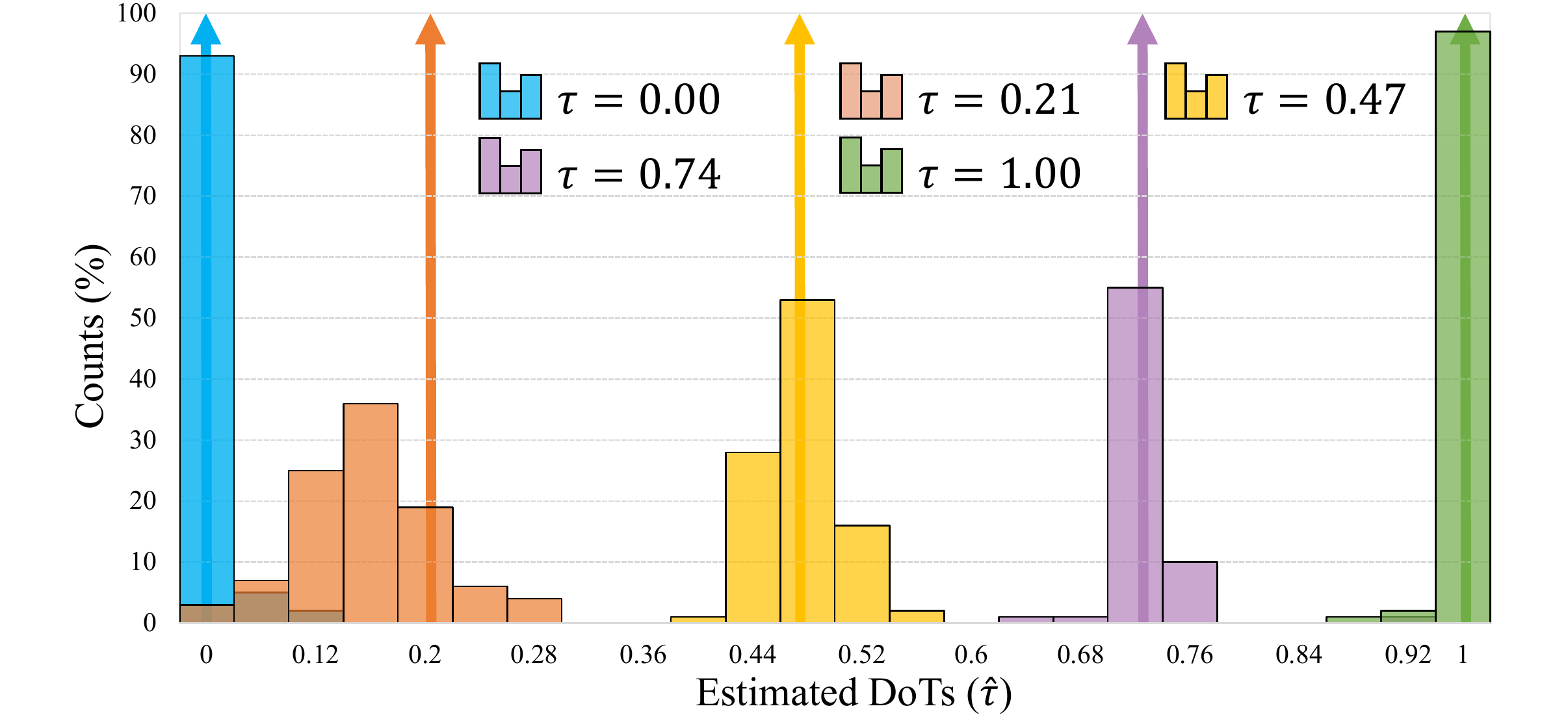}
		}\\
		\subfloat[GradCAM responses of DoT: (left) low-level responses of $\mathcal{N}^\tau$ and (right) high-level responses of $\mathcal{B}^\tau$]{
			\begin{minipage}[b]{0.38\linewidth}
				\includegraphics[width=1\linewidth]{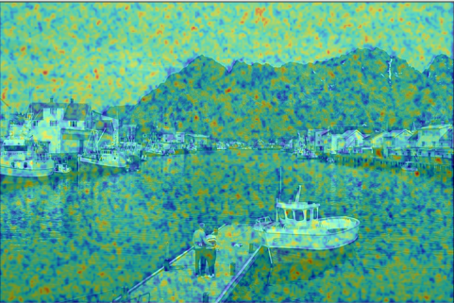}
			\end{minipage}
			\begin{minipage}[b]{0.38\linewidth}
				\includegraphics[width=1\linewidth]{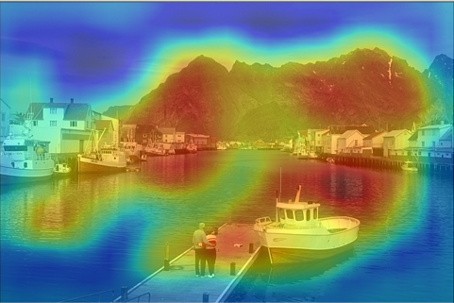}
			\end{minipage}
			\begin{minipage}[b]{0.036\linewidth}
				\includegraphics[width=1\linewidth]{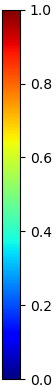}
			\end{minipage}
		}
		\caption{Statistical distribution and visual explanation of DoT estimation.}
		\label{fig:transitional distribution}
	\end{figure*}
	
	\subsection{Implementation Details}{\label{sec:implementations}}
	As in Fig.~\ref{fig:architecture}, we set $n=32$ residual blocks~\cite{LimB2017CVPRW} with $c=128$ channels for homogeneous feature extraction, and sample $T=8$ local patches with random cropping boxes of size $32\times32$ for DoT estimation. Furthermore, we feed the estimated DoTs and homogeneous features into the transitional learning module which stacks $m=8$ residual blocks for network interpolation, and finally use sub-pixel convolutions~\cite{ShiW2016CVPR} for feature upscaling. 
	
	As discussed in Section~\ref{sec:transitionality of degradations} and following~\cite{WangL2021CVPR}, two typical degradations are applied to generate the corresponding LR images, including convolutive degradations (isotropic Gaussian blur kernels of size $21\times21$, where $\sigma\sim\mathbb{U}(0.2, 2.0)$ for $\times2$ upscaling and $\sigma\sim\mathbb{U}(0.2, 4.0)$ for $\times4$ upscaling), and additive degradations (AWGN with noise levels $\mathcal{N}\sim\mathbb{U}(0,30)$). As illustrated in Eq.(\ref{eq:degradation model}), since the additive degradation $\mathcal{N}$ is independent of the convolutive degradation $\mathcal{B}$, we consider them separately in this paper.
	
	All models are optimized using the Adam optimizer~\cite{KingmaD2014ICLR_Adam} with a mini-batch of size $B$=16, and each mini-batch contains $B$ paired groups of $48\times48$ RGB LR patches and the corresponding $48s\times48s$ HR ground truth for $\times s$ upscaling. These LR-HR pairs are augmented with random horizontal and vertical flips and 90 rotations. All LR and HR images are preprocessed by subtracting the mean RGB value of the training sets.
	During the training, the learning rate is initialized to ${2\times{10}^{-4}}$ and halved for every $10^5$ mini-batch update, then the final models will reach convergence after $3\times10^5$ iterations. The experiments are conducted in PyTorch framework with a 20-core Intel Xeon Gold CPU and a single 32 GB NVIDIA Tesla V100 GPU.
	\subsection{Effect of Transitional Learning}
	As illustrated in Fig.~\ref{fig:transitional_modal_learning}, the transitional learning is built on at least two primary models with $\Theta^0$ and $\Theta^1$. Remarkably, each primary model acts as a specific SISR baseline, with a specific solution space for a given degradation, and shows strong performance on its specific degradation but weak capacities on other unseen degradations.
	
	To investigate the effects of transitional learning, we conduct two baseline models without transitional learning for either additive or convolutive degradations:\\
	1) baseline-$\mathcal{D}^0$: $\boldsymbol{\hat y}=\boldsymbol{\mathcal{F}}_{m}^{0}(...\mathcal{S}(\boldsymbol{\mathcal{F}}_1^{0}(\boldsymbol{x}^{\tau};\Theta_1^{0}));\Theta_{m}^{0})$\\
	2) baseline-$\mathcal{D}^1$: $\boldsymbol{\hat y}=\boldsymbol{\mathcal{F}}_{m}^{1}(...\mathcal{S}(\boldsymbol{\mathcal{F}}_1^{1}(\boldsymbol{x}^{\tau};\Theta_1^{1}));\Theta_{m}^{1})$\\
	where $\mathcal{D}$ indicates the degradation type, \ie, $\mathcal{D}$ represents $\mathcal{B}$ for isotropic Gaussian blur, or $\mathcal{N}$ for AWGN.
	Moreover, these baseline models are trained on their corresponding degradations.
	
	From Fig.~\ref{fig:ablation_study}, each baseline performs the best on the strictly corresponding degradation but gets worse performances on other degradations. For example, baseline-$\mathcal{D}^0$ (additive degradations) achieves higher PSNR on $\mathcal{N}^\tau=0$, reaching about 4dB gains against on $\mathcal{N}^\tau=30$.
	By applying the proposed transitional learning, the final TLSR obtains substantial improvements for various degradations. Specifically, our TLSR (w/ $\tau$) closes the gap of performances between $\mathcal{N}^\tau=0$ and $\mathcal{N}^\tau=30$ from 7dB to 2.6dB.
	
	Furthermore, since the real DoTs ($\tau$) are not given in blind applications, the effects of transitional learning would depend on the accuracy of DoT estimation. Nevertheless, as only a 1-D scalar should be estimated, our DoTNet is capable enough. Although without the real DoTs, our final model TLSR with the estimated DoTs ($\hat\tau$) also shows comparable performances as the non-blind TLSR (w/ $\tau$).
	
	\subsection{Effect of DoT Estimation}\label{sec:effect_DoT}
	As illustrated above, since the DoT estimation network is designed to estimate the degree of transitionality and affects the effectiveness of the transitional learning, we then conduct a visualization experiment on the distribution of the estimated DoTs to illustrate the effect of DoTNet. As in Fig.~\ref{fig:transitional distribution}(a), even though the degradation process is acting on the local regions in a single image, we can still accurately capture the DoTs by applying the random cropping and DoTNet. Nevertheless, as we mainly feed the primary degradations as the baselines and consider the transitional degradations as transition states in the training phase, the estimated DoTs are more accurate in the primary states but somewhat discrepancy in the transition states. 
	
	Therefore, to further illustrate the effect of DoT estimation, we visualize the activation responses of DoT on the different types of degradations using GradCAM~\cite{SelvarajuR2017ICCV}. As shown in Fig.~\ref{fig:transitional distribution}(b), the low-level response of DoT on additive degradation focuses on the smoothing regions of the image, which are easier to detect the noises. Contrastively, the high-level GradCAM responses of DoTs on convolutive degradations focus on the textures, which are easier to detect the blur effects. Particularly, these responses are consistent to the mechanism of human vision system, that noises in the smooth region and blurry effects in the complex textures have higher saliencies for visual perception.
	
	\begin{table*}[!t]
		\renewcommand\arraystretch{1.1}
		\centering
		\caption{Quantitative comparisons of the proposed TLSR and SOTAs on blind SR with additive degradations for $\times4$ upscaling. Particularly, VDN~\cite{YueZ2019NeurIPS} is a blind noise modelling method, we then feed the denoised image into RCAN~\cite{ZhangY2018ECCV} for SR, or use only the S-Net~\cite{YueZ2019NeurIPS} of VDN to estimate the noises for post-processing of SR using either ZSSR~\cite{ShocherA2018CVPR} or SRMD~\cite{ZhangK2018CVPR}. The inference time (sec) is averagely calculated on Set14 dataset.}
		\small
		\resizebox{1\textwidth}{!}{
		\begin{tabular}{p{0.4cm}<{\centering} c p{3.8cm} c c c c c c c c c}
			\toprule[0.2em]
			\multirow{2}{*}{Scale}&\multirow{2}{*}{$\mathcal{D}$}&\multirow{2}{*}{Method}&{Time$\downarrow$}&\multicolumn{2}{c}{Set5}&\multicolumn{2}{c}{Set14}&\multicolumn{2}{c}{BSD100}&\multicolumn{2}{c}{Urban100}\\
			&&&(sec)&PSNR$\uparrow$&SSIM$\uparrow$&PSNR$\uparrow$&SSIM$\uparrow$
			&PSNR$\uparrow$&SSIM$\uparrow$&PSNR$\uparrow$&SSIM$\uparrow$\\
			\toprule[0.1em]
			\multirow{10}{*}{$\times2$}&
			\multirow{10}{*}{\rotatebox{90}{\tabincell{c}{$\mathcal{B}^\tau=0.2$,\quad$\mathcal{N}^\tau\sim\mathbb{U}(0,30)$}}}
			&Bicubic&-
			&29.28&0.7437&27.44&0.7034&26.92&0.6749&25.11&0.6866\\
			&&RCAN~\cite{ZhangY2018ECCV}&0.139
			&27.38&0.6027&26.11&0.5981&25.47&0.5786&22.20&0.5286\\
			&&DIP~\cite{UlyanovD2018CVPR}&143.6
			&28.71&0.7911&26.45&0.6862&26.20&0.6599&25.02&0.6698\\
			&&ZSSR~\cite{ShocherA2018CVPR}&82.15
			&27.62&0.6265&26.17&0.6163&25.56&0.5937&24.31&0.6158\\
			&&VDN~\cite{YueZ2019NeurIPS}+ZSSR~\cite{ShocherA2018CVPR}&82.16
			&31.19&0.8659&29.15&0.8065&27.92&0.7568&27.40&0.8174\\
			&&VDN~\cite{YueZ2019NeurIPS}+MZSR~\cite{SohJW2020CVPR}&0.744
			&31.10&0.8177&28.80&0.7664&27.95&0.7390&27.41&0.7865\\
			&&S-Net~\cite{YueZ2019NeurIPS}+SRMD~\cite{ZhangK2018CVPR}&\textbf{0.023}
			&32.40&0.8782&29.82&0.8039&28.80&0.7649&28.17&0.8327\\
			&&DnCNN~\cite{ZhangK2017TIP}+RCAN~\cite{ZhangY2018ECCV}&0.146
			&32.94&0.9059&30.25&0.8344&29.42&0.8068&28.72&0.8515\\
			&&VDN~\cite{YueZ2019NeurIPS}+RCAN~\cite{ZhangY2018ECCV}&0.148
			&\underline{33.39}&\underline{0.9029}&\underline{30.69}&\underline{0.8443}&\underline{29.59}&\underline{0.8137}&\underline{29.39}&\underline{0.8679}\\
			&&TLSR (ours)&\underline{0.039}
			&\bf{33.67}&\bf{0.9063}&\bf{30.95}&\bf{0.8488}&\bf{29.73}&\bf{0.8166}&\bf{29.77}&\bf{0.8770}\\
			\toprule[0.1em]
			\multirow{11}{*}{$\times4$}&
			\multirow{11}{*}{\rotatebox{90}{\tabincell{c}{$\mathcal{B}^\tau=0.2$,\quad$\mathcal{N}^\tau\sim\mathbb{U}(0,30)$}}}
			&Bicubic&-
			&26.22&0.6904&24.58&0.5998&24.45&0.5614&22.23&0.5606\\
			&&RCAN~\cite{ZhangY2018ECCV}&0.086
			&24.90&0.5798&23.62&0.5221&23.09&0.4824&22.20&0.5286\\
			&&DIP~\cite{UlyanovD2018CVPR}&144.2
			&26.55&0.7280&24.68&0.6095&24.51&0.5713&22.60&0.5648\\
			&&ZSSR~\cite{ShocherA2018CVPR}&88.06
			&25.48&0.6183&23.81&0.5319&23.54&0.4857&22.49&0.5111\\
			&&VDN~\cite{YueZ2019NeurIPS}+ZSSR~\cite{ShocherA2018CVPR}&88.11
			&28.29&0.7918&26.13&0.6932&25.72&0.6512&23.84&0.6899\\
			&&VDN~\cite{YueZ2019NeurIPS}+MZSR~\cite{SohJW2020CVPR}&0.578
			&28.21&0.7975&26.18&0.6927&25.76&0.6511&23.68&0.6811\\
			&&S-Net~\cite{YueZ2019NeurIPS}+SRMD~\cite{ZhangK2018CVPR}&\textbf{0.009}
			&28.45&0.8049&26.20&0.6831&25.64&0.6348&23.94&0.6862\\
			&&DnCNN~\cite{ZhangK2017TIP}+RCAN~\cite{ZhangY2018ECCV}&0.094
			&28.62&0.8105&26.41&0.6989&25.96&0.6562&24.21&0.6994\\
			&&VDN~\cite{YueZ2019NeurIPS}+RCAN~\cite{ZhangY2018ECCV}&0.092
			&\underline{29.07}&0.8238&\underline{26.73}&0.7118&\underline{26.09}&0.6646&\underline{24.63}&\underline{0.7188}\\
			&&DASR~\cite{WangL2021CVPR}&0.033
			&28.73&\underline{0.8252}&26.55&\underline{0.7122}&25.99&\underline{0.6648}&24.16&0.7111\\
			&&TLSR (ours)&\underline{0.025}
			&\bf{29.42}&\bf{0.8349}&\bf{27.01}&\bf{0.7216}&\bf{26.22}&\bf{0.6706}&\bf{24.91}&\bf{0.7348}\\
			\bottomrule[0.2em]
		\end{tabular}
		}
		\label{tab:comparisons_SOTA_Add}
	\end{table*}

	\begin{figure*}[!h]
		\centering
		\includegraphics[width=1\linewidth]{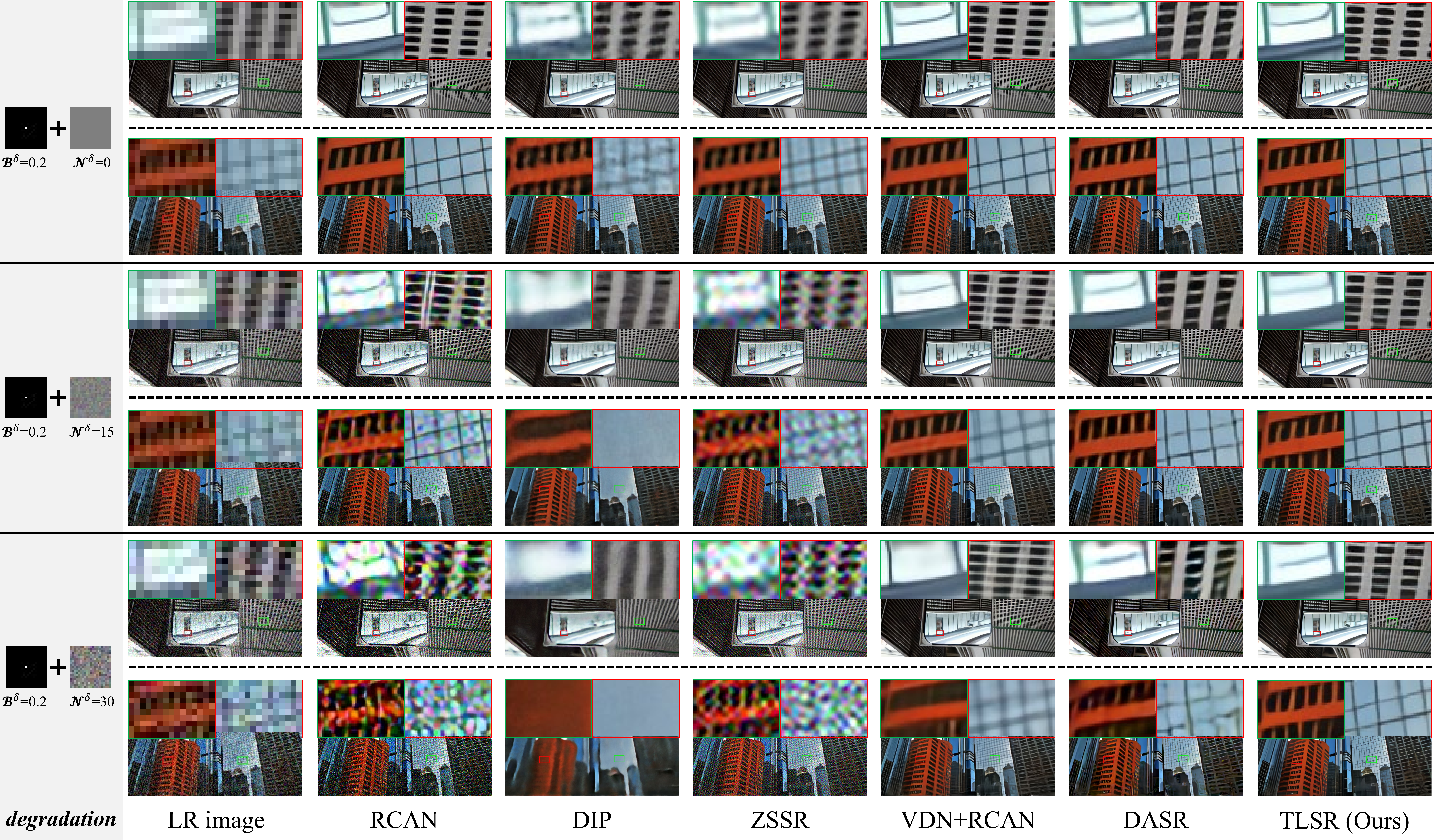}
		\caption{Qualitative visual comparisons of the proposed TLSR and state-of-the-arts blind SR methods for $\times4$ upscaling on ``{\em img004}'' and ``{\em img099}'' from Urban100 dataset with additive degradations. Our TLSR reconstructs the edges and structures better than other methods.}
		\label{fig:subjective comparisons_add}
	\end{figure*}
	
	\begin{table*}[!t]
		\renewcommand\arraystretch{1.1}
		\centering
		\caption{Quantitative comparisons of the proposed TLSR and SOTAs on blind SR with convolutive degradations for $\times4$ upscaling. Particularly, KernelGAN~\cite{BellS2019NeurIPS} is a kernel estimation method for blind image deblurring, so we apply either ZSSR~\cite{ShocherA2018CVPR} or SRMD~\cite{ZhangK2018CVPR} with the estimated kernel for post-processing of SR. The inference time (sec) is averagely calculated on Set14 dataset.}
		\small
		\resizebox{1\textwidth}{!}{
		\begin{tabular}{p{0.4cm}<{\centering} c p{3.8cm} c c c c c c c c c}
			\toprule[0.2em]
			\multirow{2}{*}{Scale}&\multirow{2}{*}{$\mathcal{D}$}&\multirow{2}{*}{Method}&\multirow{2}{*}{Time$\downarrow$}&\multicolumn{2}{c}{Set5}&\multicolumn{2}{c}{Set14}&\multicolumn{2}{c}{BSD100}&\multicolumn{2}{c}{Urban100}\\
			&&&&PSNR$\uparrow$&SSIM$\uparrow$&PSNR$\uparrow$&SSIM$\uparrow$
			&PSNR$\uparrow$&SSIM$\uparrow$&PSNR$\uparrow$&SSIM$\uparrow$\\
			\toprule[0.1em]
			\multirow{10}{*}{$\times2$}
			&\multirow{10}{*}{\rotatebox{90}{\tabincell{c}{$\mathcal{N}^\tau=0$,\quad$\mathcal{B}^\tau\sim\mathbb{U}(0.2,2.0)$}}}
			&Bicubic&-
			&30.95&0.8778&28.21&0.7954&27.79&0.7633&25.04&0.7592\\
			&&RCAN~\cite{ZhangY2018ECCV}&0.086
			&33.09&0.9001&30.03&0.8307&29.22&0.8019&27.68&0.8155\\
			&&DIP~\cite{UlyanovD2018CVPR}&143.6
			&31.70&0.8782&28.42&0.7768&27.90&0.7486&25.04&0.7197\\
			&&ZSSR~\cite{ShocherA2018CVPR}&82.15
			&32.52&0.8960&29.45&0.8232&28.60&0.7923&26.03&0.7901\\
			&&KernelGAN~\cite{BellS2019NeurIPS}+ZSSR~\cite{ShocherA2018CVPR}&374.9
			&33.64&0.9301&29.95&0.8615&29.32&0.8403&27.31&0.8478\\
			&&KernelGAN~\cite{BellS2019NeurIPS}+MZSR~\cite{SohJW2020CVPR}&293.5
			&34.03&0.9252&30.48&0.8597&29.64&0.8328&27.15&0.8345\\
			&&KernelGAN~\cite{BellS2019NeurIPS}+SRMD~\cite{ZhangK2018CVPR}&292.7
			&33.12&0.9244&29.84&0.8680&28.99&0.8449&27.11&0.8456\\
			&&IKC~\cite{GuJ2019CVPR}&0.428
			&36.38&0.9471&32.39&0.8923&31.05&0.8674&29.63&0.8833\\
			&&DASR~\cite{WangL2021CVPR}&\underline{0.043}
			&\underline{36.80}&\underline{0.9492}&\underline{32.49}&\underline{0.8934}&\underline{31.23}&\underline{0.8710}&\underline{30.18}&\underline{0.8978}\\
			&&TLSR (ours)&\textbf{0.039}
			&\bf{37.23}&\bf{0.9520}&\bf{33.08}&\bf{0.9019}&\bf{31.57}&\bf{0.8793}&\bf{30.95}&\bf{0.9092}\\
			\toprule[0.1em]
			\multirow{10}{*}{$\times4$}
			&\multirow{10}{*}{\rotatebox{90}{\tabincell{c}{$\mathcal{N}^\tau=0$,\quad$\mathcal{B}^\tau\sim\mathbb{U}(0.2,4.0)$}}}
			&Bicubic&-
			&26.43&0.7460&24.68&0.6428&24.88&0.6116&21.99&0.5938\\
			&&RCAN~\cite{ZhangY2018ECCV}&0.071
			&28.33&0.7971&26.01&0.6892&25.81&0.6526&23.55&0.6659\\
			&&DIP~\cite{UlyanovD2018CVPR}&144.2
			&27.32&0.7597&25.27&0.6471&25.19&0.6125&22.48&0.5963\\
			&&ZSSR~\cite{ShocherA2018CVPR}&88.06
			&27.35&0.7747&25.31&0.6683&25.34&0.6353&22.66&0.6236\\
			&&KernelGAN~\cite{BellS2019NeurIPS}+ZSSR~\cite{ShocherA2018CVPR}&376.1
			&28.61&0.8093&26.47&0.7031&26.03&0.6802&23.69&0.6823\\
			&&KernelGAN~\cite{BellS2019NeurIPS}+MZSR~\cite{SohJW2020CVPR}&294.2
			&28.59&0.8112&26.39&0.7083&25.99&0.6768&23.71&0.6795\\
			&&KernelGAN~\cite{BellS2019NeurIPS}+SRMD~\cite{ZhangK2018CVPR}&292.9
			&28.16&0.8106&26.03&0.7031&25.79&0.6801&23.60&0.6761\\
			&&IKC~\cite{GuJ2019CVPR}&0.219
			&30.58&0.8579&27.85&0.7532&27.00&0.7105&24.95&0.7409\\
			&&DASR~\cite{WangL2021CVPR}&\underline{0.033}
			&\underline{31.40}&\underline{0.8790}&\underline{28.06}&\underline{0.7570}&\underline{27.22}&\underline{0.7135}&\underline{25.30}&\underline{0.7523}\\
			&&TLSR (ours)&\textbf{0.025}
			&\bf{31.62}&\bf{0.8836}&\bf{28.34}&\bf{0.7653}&\bf{27.37}&\bf{0.7203}&\bf{25.71}&\bf{0.7683}\\
			\bottomrule[0.2em]
		\end{tabular}
		}
		\label{tab:comparisons_SOTA_Conv}
	\end{table*}
	
	\begin{figure*}[!h]
		\centering
		\includegraphics[width=1\linewidth]{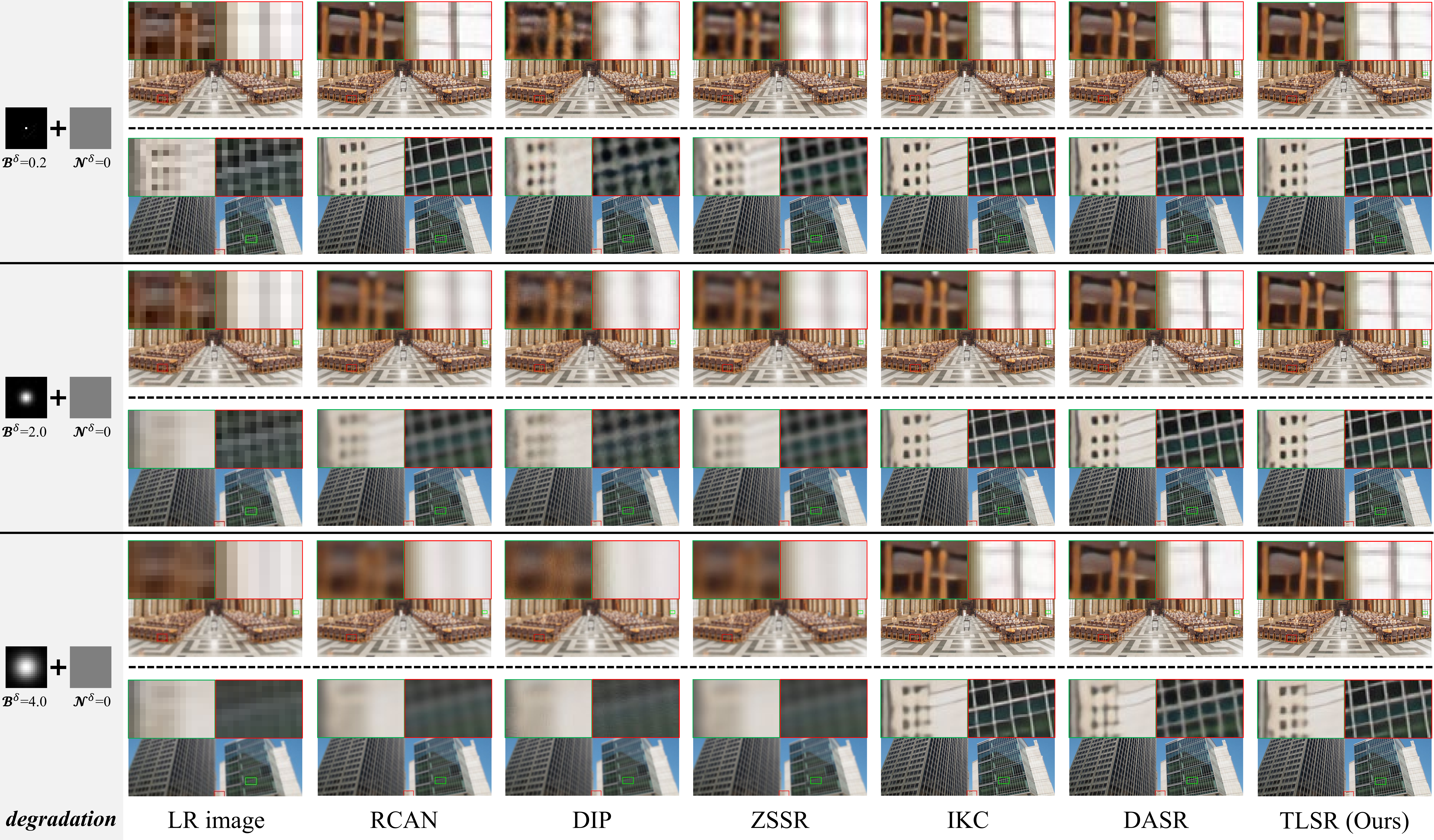}
		\caption{Qualitative comparisons of the proposed TLSR and state-of-the-arts blind SR methods for $\times4$ upscaling on ``{\em img079}'' and ``{\em img096}'' from Urban100 dataset with convolutive degradations. Our TLSR has fewer distortions and better structures/edges than other methods.}
		\label{fig:subjective comparisons_conv}
	\end{figure*}

	\begin{table}[!t]
		\renewcommand\arraystretch{1.2}
		\centering
		\small
		\caption{Quantitative comparisons of the proposed TLSR and state-of-the-art SR methods on computational complexities for $\uparrow_{\times2}$ upscaling. Particularly, the super-resolved result is a single $720\times480$ RGB image.}
		\resizebox{1.0\linewidth}{!}{
			\begin{tabular}{l c c c c}
				\toprule[0.2em]
				&RCAN&ZSSR&IKC&TLSR\\
				\toprule[0.1em]
				\textit{Params}(M)&15.44&\textbf{0.59}&4.09+0.43+0.65&5.14+0.62\\
				\textit{FLOPs}(G)&1325.9&205.1$\times$2000&$\approx$357.4$\times$8&\textbf{$\approx$320.0}\\
				\textit{Time}(sec)&0.124&86.61&0.596&\textbf{0.041}\\
				\bottomrule[0.2em]
			\end{tabular}
		}
		\label{tab:complexity}
	\end{table}

	\begin{table*}[!t]
		\renewcommand\arraystretch{1.1}
		\centering
		\small
		\caption{Quantitative comparisons (PSNR$\uparrow$) of the proposed TLSR$_\text{Real}$ and SOTAs on blind SR with mixed synthetic and unknown real-world degradations for $\times4$ upscaling.}
		\resizebox{1.0\textwidth}{!}{
			\begin{tabular}{c l c c c c c c c c c c c}
				\toprule[0.2em]
				\multirow{2}{*}{Dataset}&\multirow{2}{*}{Method}&\multirow{2}{*}{Time$\downarrow$}&\multicolumn{3}{c}{$\mathcal{B}^\tau$=$0.2$}&\multicolumn{3}{c}{$\mathcal{B}^\tau$=$1.0$}&\multicolumn{3}{c}{$\mathcal{B}^\tau$=$2.0$}\\
				&&&$\mathcal{N}^\tau$=$0$&$\mathcal{N}^\tau$=$5$&$\mathcal{N}^\tau$=$10$&$\mathcal{N}^\tau$=$0$&$\mathcal{N}^\tau$=$5$&$\mathcal{N}^\tau$=$10$&$\mathcal{N}^\tau$=$0$&$\mathcal{N}^\tau$=$5$&$\mathcal{N}^\tau$=$10$\\
				\toprule[0.1em]
				\multirow{6}{*}{Set14}
				&ZSSR~\cite{ShocherA2018CVPR}&88.23
				&27.33&26.59&25.18&26.98&26.30&24.96&25.53&25.07&24.06\\
				&DIP~\cite{UlyanovD2018CVPR}&144.2
				&27.11&26.33&25.52&26.82&26.12&25.27&25.46&24.92&24.39\\
				&DnCNN~\cite{ZhangK2017TIP}+IKC~\cite{GuJ2019CVPR}&0.221
				&27.90&27.36&26.70&28.00&27.33&26.60&27.59&26.12&25.61\\
				&VDN~\cite{YueZ2019NeurIPS}+IKC~\cite{GuJ2019CVPR}&0.227
				&\underline{28.36}&\underline{27.73}&26.95&\underline{28.40}&\underline{27.60}&26.83&\underline{27.91}&26.40&25.84\\
				&DASR~\cite{WangL2021CVPR}&\textbf{0.033}
				&28.08&27.66&\textbf{27.07}&28.15&27.58&\textbf{26.91}&27.79&\underline{26.79}&\underline{26.08}\\
				&TLSR$_\text{Real}$(Ours)&\underline{0.051}
				&\textbf{28.79}&\textbf{27.80}&\underline{27.04}&\textbf{28.73}&\textbf{27.68}&\underline{26.87}&\textbf{28.65}&\textbf{26.95}&\textbf{26.09}\\
				\hline
				\multirow{6}{*}{Urban100}
				&ZSSR~\cite{ShocherA2018CVPR}&137.3
				&24.79&24.48&23.66&24.70&24.36&23.56&23.55&23.39&22.82\\
				&DIP~\cite{UlyanovD2018CVPR}&308.1
				&24.36&23.89&23.23&24.27&23.75&23.19&23.47&22.97&22.44\\
				&DnCNN~\cite{ZhangK2017TIP}+IKC~\cite{GuJ2019CVPR}&0.593
				&25.30&24.77&24.17&25.30&24.69&24.09&24.74&23.48&23.08\\
				&VDN~\cite{YueZ2019NeurIPS}+IKC~\cite{GuJ2019CVPR}&0.602
				&25.75&25.18&24.55&25.68&25.03&24.40&24.92&23.74&23.32\\
				&DASR~\cite{WangL2021CVPR}&\textbf{0.040}
				&25.22&24.96&24.55&25.19&24.82&24.39&24.78&24.07&23.58\\
				&TLSR$_\text{Real}$(Ours)&\underline{0.075}
				&\textbf{26.38}&\textbf{25.27}&\textbf{24.62}&\textbf{26.14}&\textbf{25.07}&\textbf{24.45}&\textbf{25.94}&\textbf{24.37}&\textbf{23.69}\\
				\bottomrule[0.2em]
			\end{tabular}
		}
		\label{tab:comparisons_SOTA_Real}
	\end{table*}
	\begin{figure*}[!h]
		\centering
		\includegraphics[width=1.\linewidth]{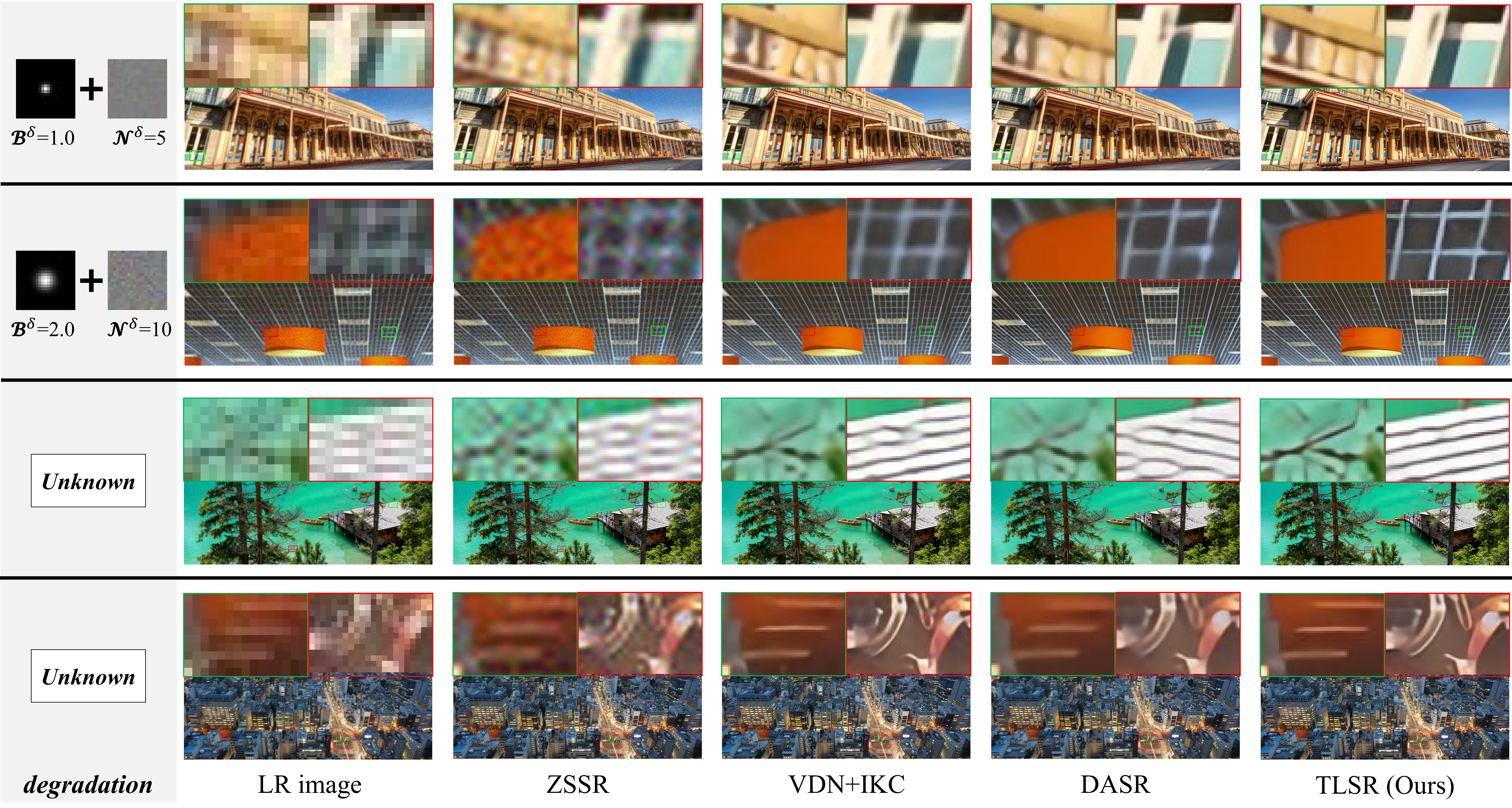}
		\caption{SR results on images with mixed synthetic degradations or unknown real-world degradations~\cite{NTIRE2020CVPRW} for $\times4$ upscaling. Specifically, from top to bottom, the images are ``{\em img017}'' and ``{\em img044}'' from Urban100 dataset, ``{\em 0807}'' and ``{\em 0873}'' from real-world DIV2K~\cite{NTIRE2020CVPRW} dataset.}
		\label{fig:real_world}
	\end{figure*}
	\subsection{Experiments on Unmixed Degradations}
	In this subsection, we evaluate quantitative and qualitative comparisons on the unmixed synthetic (additive or convolutive) degradations, and the comparative methods are \\
	1) non-blind SISR method, as RCAN~\cite{ZhangY2018ECCV};\\
	2) integrated learning based blind SR methods (IL-BSR), including DnCNN~\cite{ZhangK2017TIP}, VDN~\cite{YueZ2019NeurIPS}, IKC~\cite{GuJ2019CVPR} and DASR~\cite{WangL2021CVPR};\\
	3) zero-shot learning based blind SR methods (ZSL-BSR), including DIP~\cite{UlyanovD2018CVPR}, ZSSR~\cite{ShocherA2018CVPR}, KernelGAN~\cite{BellS2019NeurIPS} and MZSR~\cite{SohJW2020CVPR}.
	
	Specifically, VDN and KernelGAN are the representative blind methods for noise modeling and kernel estimation, respectively, and we naturally apply the ZSSR with the estimated noises or kernels for blind SR as in \cite{BellS2019NeurIPS}.
	Besides, for general blind SR: 1) IKC iteratively estimates and corrects the representation vector of blur kernel to guide a conditional integrated model for blind deblurring and super-resolution; 2) DASR directly predict a discriminative degradation representation vector to guide a conditional integrated model, which is efficient as a non-iterative blind SR method; 3) DIP is an image-specific heuristic method to progressively explore an ideal transition state between random noises and the corrupted LR images\footnote{As specified in~\cite{UlyanovD2018CVPR}, on synthetic degradations, we use the ground-truth to select the number of iterations with the best PSNR, and on unknown degradations, we fix the number of iterations to 2000.}; 4) ZSSR is a representative zero-shot method by iteratively optimizing an image-specific model using internal exemplars; 5) MZSR is a meta-learning method by adaptively correcting a pre-trained model to solve image-specific blind SR task. Particularly, since the ZSL-BSR models are image-specific and trained online, we use the released codes from the authors for training and inference. 
	
	For evaluation, the LR images are synthesized with variational degradations whose DoTs are set range from 0 to 1, namely, $\mathcal{N}^\tau\in\{0,5,10,15,20,25,30\}$ for both $\times2$ and $\times4$ upscaling, $\mathcal{B}^\tau\in\{0.2,0.5,1.0,1.5,2.0\}$ for $\times2$ upscaling, and $\mathcal{B}^\tau\in\{0.2,1.0,2.0,3.0,4.0\}$ for $\times4$ upscaling. And the quantitative results are averaged in the corresponding multiple degradations.
	
	Ascribing to the vital capacity of transitional learning to rebuild transitional networks, TLSR is more flexible and capable of processing various degradations as long as the degradations are transitional. Therefore, as reported in TABLE~\ref{tab:comparisons_SOTA_Add} and TABLE~\ref{tab:comparisons_SOTA_Conv}, without any additional iterative operations for correcting $\hat\tau$, our TLSR achieves best performances on both additive and convolutive degradations, and is faster than most of the comparative SOTA methods. For example, for $\times2$ upscaling on Set14 dataset with convolutive degradations, our TLSR gains PSNR of 0.69dB and 3.63dB, and is $\times10$ and $\times2000$ times faster than IKC and ZSSR, respectively. Moreover, from the detailed quantitative comparison on computational complexities reported in TABLE~\ref{tab:complexity}, our TLSR is with relatively the fewest complexities and fastest speed.
	
	Besides, the qualitative comparisons (for $\times4$ upscaling) are shown in Fig.~\ref{fig:subjective comparisons_add} and Fig.\ref{fig:subjective comparisons_conv}. Obviously, for the additive degradations $\mathcal{N}\in\{0,15,30\}$, our TLSR reconstructs the noisy edges and structures better than other methods, especially for $\mathcal{N}\ge15$, the existing state-of-the-art blind SR methods tend to generate more distortions and smoothen the details. For the convolutive degradations $\mathcal{B}\in\{0.2,2.0,4.0\}$, our TLSR shows higher fidelity of color and details against other blind SR methods, particularly limited by the iterative operations, IKC shows comparative performance but are far slower than our TLSR.
	
	
	\subsection{Experiments on Mixed Degradations}
	
	Besides the unmixed synthetic degradations, we further conduct experiments on more complex (real-world) degradations.
	Since the degradation of the real-world image is generally unknown and mixed, but our TLSR models are trained on a single type of transitional degradations (\eg, $\boldsymbol{\mathcal{F}}_{\text{TLSR}_\mathcal{N}}$ for additive degradations, and or $\boldsymbol{\mathcal{F}}_{\text{TLSR}_\mathcal{B}}$ for convolutive degradations). As in Eq.(\ref{eq:degradation model}) and following~\cite{WangL2021CVPR}, since the additive degradation $\mathcal{N}$ is independent of the convolutive degradation $\mathcal{B}$, it is feasible to plug-in the denoiser as a preprocessing module. 
	
	Therefore, we build a model as $\text{TLSR}_{\text{Real}}$ by firstly using $\text{TLSR}_\mathcal{N}$ for denoising and then feeding the denoised image $\boldsymbol{x'}$ into the $\text{TLSR}_\mathcal{B}$ for deblurring and super-resolution:
	
	\begin{equation}
		\begin{aligned}
			\boldsymbol{x'} &= \boldsymbol{\mathcal{F}}_{\text{TLSR}_{\mathcal{N}}}(\boldsymbol{x};\Theta^{\boldsymbol{\mathcal{F}}_{\text{DoT}_{\mathcal{N}}}(\boldsymbol{x})}, \uparrow_{\times1}) \\
			\boldsymbol{y} &= \boldsymbol{\mathcal{F}}_{\text{TLSR}_{\mathcal{B}}}(\boldsymbol{x'};\Theta^{\boldsymbol{\mathcal{F}}_{\text{DoT}_{\mathcal{B}}}(\boldsymbol{x'})},\uparrow_{\times4})
		\end{aligned}
		\label{eq:TLSR_Real}
	\end{equation}
	additionally, as illustrated in Section~\ref{sec:effect_DoT}, the deblurring and denoising tasks have opposite effects on DoT representation. Thus, it is feasible to solve the complex degradations using this two-stage framework. 
	
	As in TABLE~\ref{tab:comparisons_SOTA_Real} and Fig.~\ref{fig:real_world}, ascribing to the flexible transitional learning, on both complex degradations and unknown real-world degradations (from {\em NTIRE 2020 Challenge in Real-World SR}~\cite{NTIRE2020CVPRW}), our $\text{TLSR}_{\text{Real}}$ achieves comparable performances against other blind SR methods.
	Nevertheless, as a two-stage framework, TLSR$_\text{Real}$ costs twice of a single TLSR framework, and it is promising to solve DoT estimation on complex degradations for implementing one-stage TLSR$_\text{Real}$.
	
	\section{Discussions}
	Apart from the aforementioned additive and convolutive degradations, other types of degradation also exist in realistic scenes. In this section, we mainly investigate the intermediate variants of a single type of degradation experimentally, but the corresponding theoretical demonstrations are expected for further exploring. 	
	\begin{table}[!t]
		\renewcommand\arraystretch{1.1}
		\centering
		\small
		\caption{Quantitative comparisons (PSNR$\uparrow$) of the proposed TLSR and SOTAs on blind SR with anisotropic Gaussian blur degradations for $\times4$ upscaling on Urban100 dataset.}
			\begin{tabular}{p{2.6cm} c c c c c}
				\toprule[0.2em]
				\multirow{2}{*}{Method}&\multicolumn{5}{c}{Anisotropic Blur Kernel}\\
				&\includegraphics[width=0.5cm, height=0.5cm]{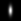}
				&\includegraphics[width=0.5cm, height=0.5cm]{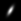}
				&\includegraphics[width=0.5cm, height=0.5cm]{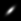}
				&\includegraphics[width=0.5cm, height=0.5cm]{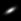}
				&\includegraphics[width=0.5cm, height=0.5cm]{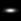}\\
				\toprule[0.1em]
				Bicubic&21.76&21.74&21.74&21.74&21.75\\
				ZSSR~\cite{ShocherA2018CVPR}
				&23.22&23.21&23.17&23.11&23.06\\
				DIP~\cite{UlyanovD2018CVPR}
				&23.11&23.09&23.05&23.03&23.09\\
				IKC~\cite{GuJ2019CVPR}&23.03&23.32&23.19&23.30&23.19\\
				DASR~\cite{WangL2021CVPR}&\underline{24.60}&\underline{24.63}&\underline{24.67}&\underline{24.68}&\underline{24.64}\\
				TLSR (Ours)&\bf{25.84}&\bf{25.76}&\bf{25.79}&\bf{25.83}&\bf{25.84}\\
				\bottomrule[0.2em]
			\end{tabular}
		\label{tab:comparisons_AnisoGauss}
	\end{table}
	\subsection{Anisotropic Gaussian Blur}
	Different from the isotropic Gaussian kernel, anisotropic Gaussian kernel is more flexible to represent the blur effects by a linear moving camera, which is formulated as $\mathcal{B}=\mathbb{N}(0, \Sigma)$. Unlike a single scalar $\sigma$ for representing isotropic Gaussian kernels, $\Sigma$ indicates a covariance matrix determined by two scalars $[\sigma_u,\sigma_v]$ and a rotation angle $\theta$. Since $\Sigma$ has higher degree of freedom and is too complicated for the present TLSR with a single DoT, we then fix $\sigma_u=1.3$, $\sigma_v=3.25$ and randomly sample the rotation angles $\theta\sim\mathbb{U}(0, \pi/2)$. The other implementation details are the same as Section~\ref{sec:implementations}.
	
\begin{figure*}[!t]
		\centering
		\includegraphics[width=1\linewidth]{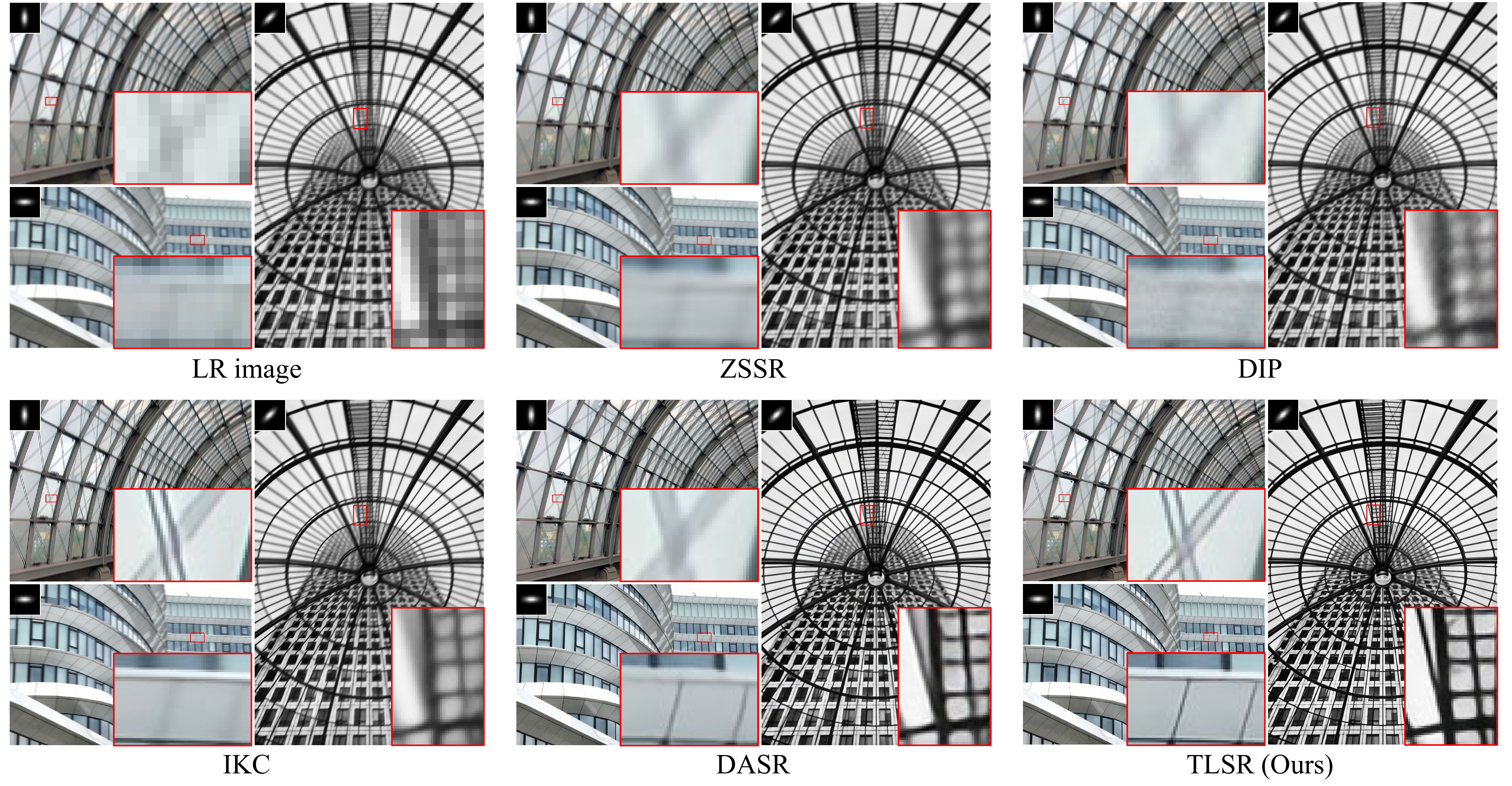}
		\caption{Qualitative comparisons of the proposed TLSR and state-of-the-arts blind SR methods on ``{\em img002}'', ``{\em img052}'' and ``{\em img072}'' from Urban100 dataset for $\times4$ upscaling with different anisotropic Gaussian blur kernels.}
		\label{fig:aniso}
	\end{figure*}

	As reported in TABLE~\ref{tab:comparisons_AnisoGauss}, unlike IKC and DASR that estimate a high-dimension representation vector to distinguish	different degradations, our TLSR explores only a 1-D scalar $\tau$ to represent the different rotation angles, which is easier to train and achieves the best performance on the Urban100 dataset with $\theta\in\{0,\pi/6,\pi/4,\pi/3,\pi/2\}$. For qualitative comparisons, we post the visual comparisons as shown in Fig.~\ref{fig:aniso}, it is observed that the degraded LR images suffer from seriously motion blurry effects in $v$-axis, and only our TLSR could restore them well for all the rotation angles.
	\begin{figure}[!t]
		\centering
		\includegraphics[width=1\linewidth]{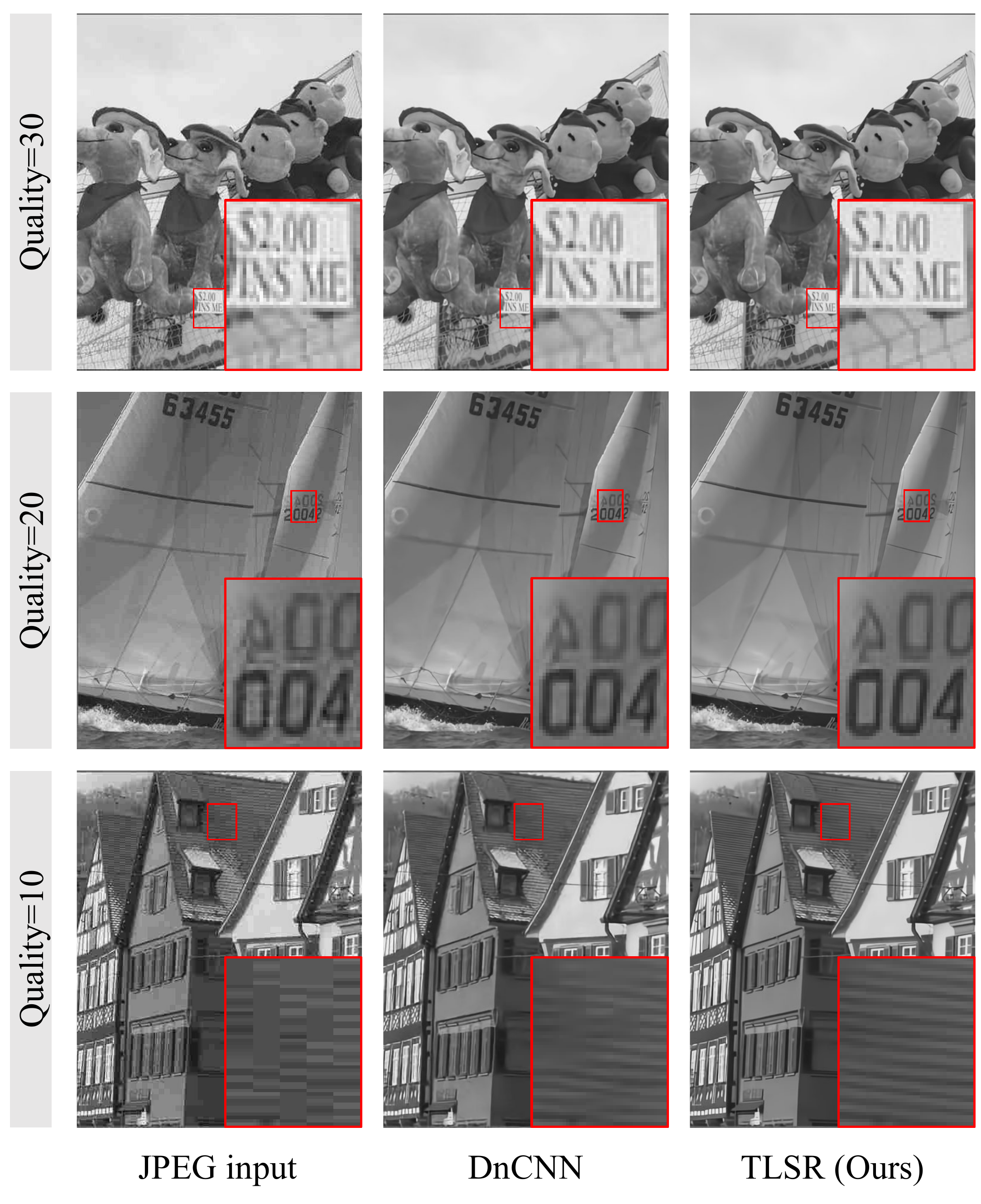}
		\caption{Qualitative comparisons of the proposed TLSR and SOTAs on ``{\em buildings}'', ``{\em sailing3}'' and ``{\em carnivaldolls}'' from LIVE1 dataset for JPEG compression artifacts removal with $Q$=10, $Q$=20 and $Q$=30, respectively.}
		\label{fig:JPEG}
		\vspace{-0.2cm}
	\end{figure}

	\begin{table}[!t]
		\renewcommand\arraystretch{1.1}
		\centering
		\small
		\caption{Quantitative comparisons (PSNR$\uparrow$) of the proposed TLSR and SOTAs on blind JPEG compession artifacts removal.}
			\begin{tabular}{p{1.7cm} p{0.7cm}<{\centering} p{0.7cm}<{\centering} p{0.7cm}<{\centering} p{0.7cm}<{\centering} p{0.7cm}<{\centering} p{0.7cm}<{\centering} p{0.7cm}<{\centering} p{0.7cm}<{\centering} p{0.7cm}<{\centering}}
				\toprule[0.2em]
				\multirow{2}{*}{Method}&\multicolumn{3}{c}{Classic5}&\multicolumn{3}{c}{LIVE1}\\
				&$Q$=10&$Q$=20&$Q$=30&$Q$=10&$Q$=20&$Q$=30\\
				\toprule[0.1em]
				JPEG&27.82&30.12&31.48&27.77&30.07&31.41\\
				LD~\cite{LiY2014ECCV}&28.39&30.30&31.50&28.26&30.19&31.32\\
				ARCNN~\cite{DongC2015CVPR}&29.04&31.16&32.52&28.98&31.29&32.69\\
				TNRD~\cite{ChenY2017TPAMI}&29.28&31.47&32.78&29.15&31.46&32.84\\
				DnCNN~\cite{ZhangK2017TIP}&29.40&31.63&32.91&29.19&31.59&32.98\\
				DCSC~\cite{FuX2019ICCV}&29.25&31.43&32.68&29.17&31.48&32.83\\
				QGAC~\cite{EhrlichM2020ECCV}&\underline{29.84}&\underline{31.98}&\underline{33.22}&\underline{29.53}&\underline{31.86}&\underline{33.23}\\
				TLSR(Ours)&\bf{29.92}&\bf{32.07}&\bf{33.33}&\bf{29.62}&\bf{31.99}&\bf{33.41}\\
				\bottomrule[0.2em]
			\end{tabular}
		\label{tab:comparisons_JPEG}
	\end{table}
	\subsection{JPEG Compression Artifacts}
	In practice, especially social media, JPEG compression is utilized to speed up image transmission, leading to the loss of high-frequency details. Fortunately, since JPEG compression is based on the non-overlapping $8\times8$ pixel blocks and is determined by a single scalar of quality factor $Q$, the DoT estimation with random cropping boxes of size $\min(width,height)\ge8$ is appropriate enough. Thus, same as Section~\ref{sec:implementations}, we also set the box size to $32\times32$ but remove the sub-pixel convolutions for upscaling, and random sample the quality factors $Q\sim\mathbb{U}(10,30)$ to train a TLSR model for blind compression artifacts removal.
	
	As reported in TABLE~\ref{tab:comparisons_JPEG}, our TLSR achieves the best performance on both Classic5 and LIVE1 datasets for various JPEG compression artifacts removal. For qualitative comparisons, we post the visual comparisons as shown in Fig.~\ref{fig:JPEG}, it is observed that the compressed images suffer from seriously smoothing in each $8\times8$ blocks. After applying our TLSR, both the flattened regions and textures are restored well.
	
	
	\section{Conclusion}
	In this paper, we firstly analyze and demonstrate the transitionality of degradations theoretically, which indicates that any transitional degradation can be well represented by two or more primary degradations. Under this condition, we then propose a transitional learning method for blind super-resolution on both additive degradation (AWGN) and convolutive degradations (isotropic Gaussian blur kernel). Specifically, the SR performance and generalization are significantly improved by introducing the DoT estimation and constructing an adaptively transitional transformation function using transitional learning under the guidance of the estimated DoT. Theoretical and experimental demonstrations indicate that the proposed TLSR achieves superior performance on synthetic unmixed/mixed degradations and unknown real-world degradation, and consumes relatively less time against the state-of-the-art blind SR methods. Furthermore, we also discuss the performance of our TLSR on blind SR with anisotropic Gaussian blur kernel and blind JPEG compression artifacts removal tasks, but the corresponding theoretical demonstrations on the transitionality of these degradations are expected for further exploring.
	
	\bibliographystyle{IEEEtran}
	\bibliography{TLSR}
	\vspace{-0.4cm}
	\begin{IEEEbiography}[{\includegraphics[width=1in,height=1.25in,clip,keepaspectratio]{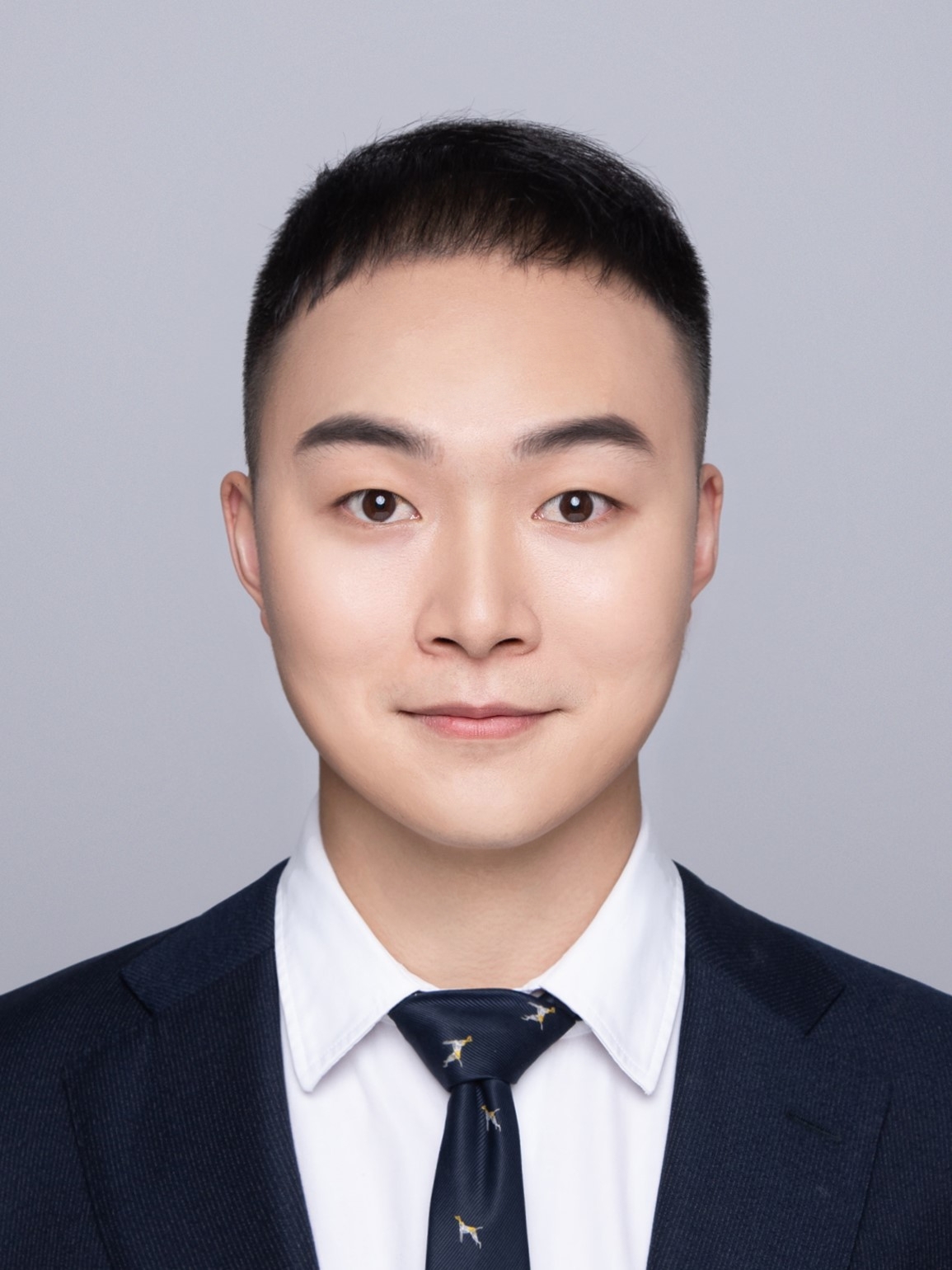}}]{Yuanfei Huang} received the B.Eng. degree in communication engineering from Hebei University of Technology, Tianjin, China, in 2016, and the Ph.D. degree in circuits and systems from the School of Electronic Engineering, Xidian University, Xi'an, China, in 2021. He is currently a Post-Doctoral Research Fellow with the School of Artificial Intelligence, Beijing Normal University, Beijing, China. His current research interests include low-level vision, computational imaging, and machine learning. In these areas, he has published technical articles in referred journals and proceedings including IEEE TIP, IEEE TCSVT, Signal Processing, \etc.
	\end{IEEEbiography}
	
	\begin{IEEEbiography}[{\includegraphics[width=1in,height=1.25in,clip,keepaspectratio]{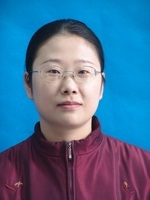}}]{Jie Li} received the B.Sc., M.Sc. and Ph.D. degrees in Circuits and Systems from Xidian University, China, in 1995, 1998 and 2005 respectively. Since 1998, she joined the School of Electronic Engineering at Xidian University. Currently, she is a Professor of Xidian University. Her research interests include computational intelligence, machine learning, and image processing. In these areas, she has published over 50 technical articles in refereed journals and proceedings including IEEE TIP, IEEE TCyb, IEEE TCSVT, Pattern Recognition, \etc.
	\end{IEEEbiography}
	
	\begin{IEEEbiography}[{\includegraphics[width=1in,height=1.25in,clip,keepaspectratio]{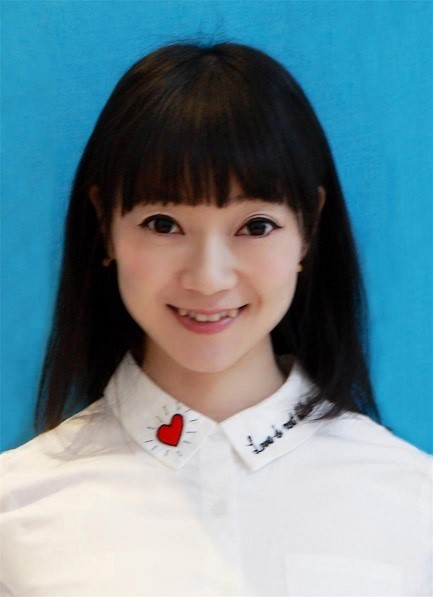}}]{Yanting Hu} received the M.Eng. degree in information and communication engineering and Ph.D. degree in pattern recognition and intelligent system from Xidian University, Xi'an, China, in 2008 and 2019, respectively. Since 2008, she has been a faculty member with the School of Medical Engineering and Technology, Xinjiang Medical University, Urumqi, China. Her current research interests include machine learning and computer vision.
	\end{IEEEbiography}
	
	\begin{IEEEbiography}[{\includegraphics[width=1in,height=1.25in,clip,keepaspectratio]{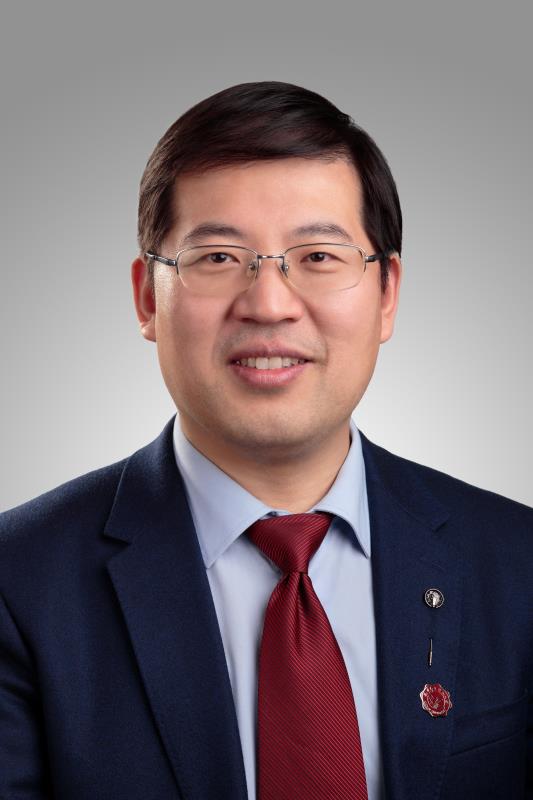}}]{Xinbo Gao}(M'02-SM'07) received the B.Eng., M.Sc. and Ph.D. degrees in electronic engineering, signal and information processing from Xidian University, Xi'an, China, in 1994, 1997, and 1999, respectively. From 1997 to 1998, he was a research fellow at the Department of Computer Science, Shizuoka University, Shizuoka, Japan. From 2000 to 2001, he was a post-doctoral research fellow at the Department of Information Engineering, the Chinese University of Hong Kong, Hong Kong. Since 2001, he has been at the School of Electronic Engineering, Xidian University. He is a Cheung Kong Professor of Ministry of Education of P. R. China, a Professor of Pattern Recognition and Intelligent System of Xidian University. Since 2020, he is also a Professor of Computer Science and Technology of Chongqing University of Posts and Telecommunications. His current research interests include Image processing, computer vision, multimedia analysis, machine learning and pattern recognition. He has published six books and around 300 technical articles in refereed journals and proceedings. Prof. Gao is on the Editorial Boards of several journals, including Signal Processing (Elsevier) and Neurocomputing (Elsevier). He served as the General Chair/Co-Chair, Program Committee Chair/Co-Chair, or PC Member for around 30 major international conferences. He is a Fellow of the Institute of Engineering and Technology and a Fellow of the Chinese Institute of Electronics.
	\end{IEEEbiography}
	
	\begin{IEEEbiography}[{\includegraphics[width=1in,height=1.25in,clip,keepaspectratio]{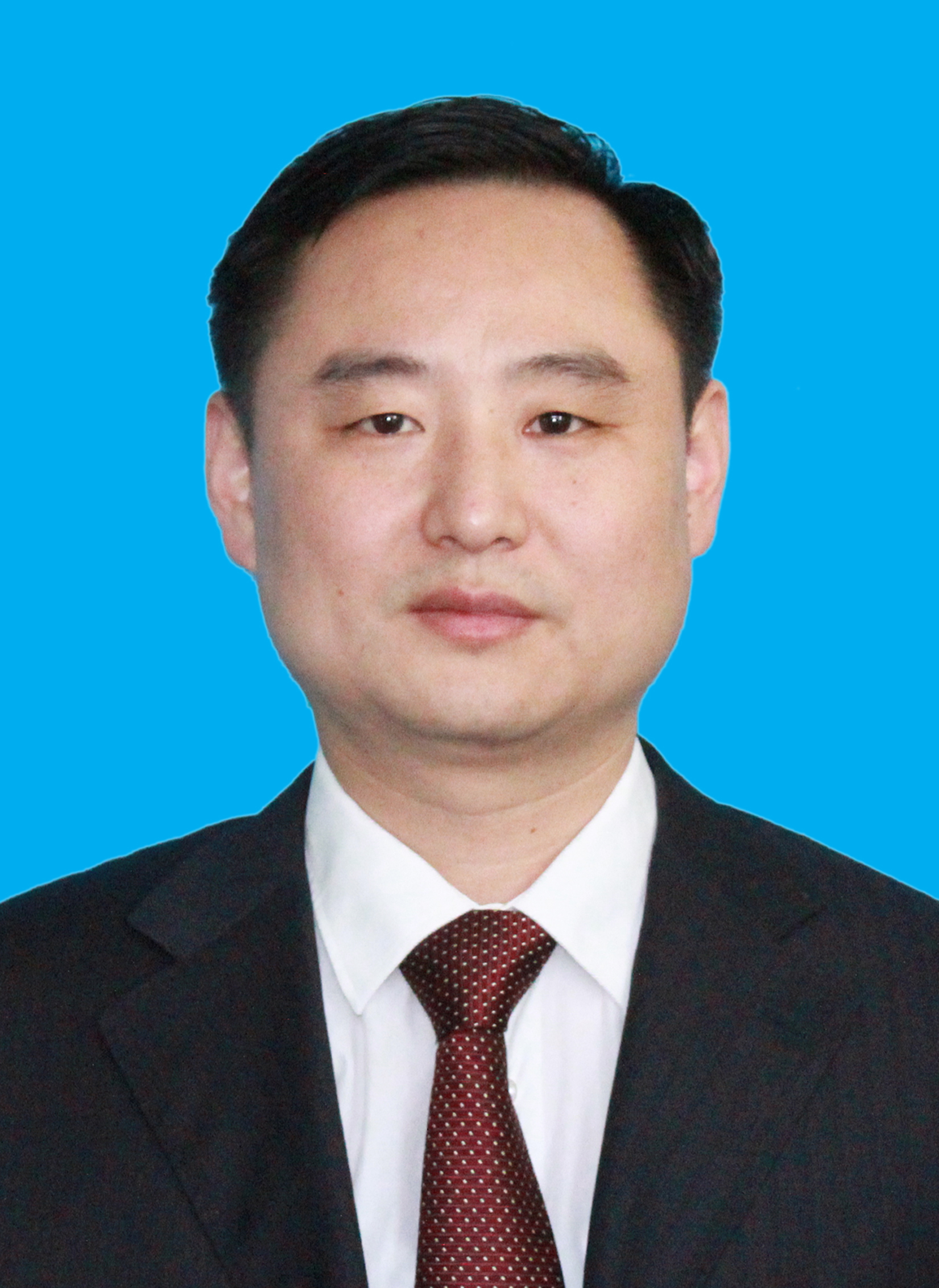}}]{Hua Huang}(Senior Member, IEEE) received his B.S., M.S., and Ph.D. degrees from Xi’an Jiaotong University, Xian, China, in 1996, 2001, and 2006, respectively. He is currently a Professor in School of Artificial Intelligence, Beijing Normal University. His current research interests include image and video processing, computer graphics, and pattern recognition.
	\end{IEEEbiography}

	\vfill
	\clearpage
	\onecolumn
	\setcounter{section}{0}
	\setcounter{proposition}{0}
	\setcounter{MYtempeqncnt}{\value{equation}}
	\setcounter{equation}{0}
	\section{Appendix}
	
	\subsection{Proof of Proposition 1}{\label{appendix:proof}}
	\begin{proposition}
		Given two Gaussian blur kernels $\mathcal{B}^0\sim\mathbb{N}(0, \sigma^2_0)$, $\mathcal{B}^1\sim\mathbb{N}(0, \sigma^2_1)$ ($\sigma_1>\sigma_0$), and degree of transitionality $\tau$, then in element-wise $(i,j)$, $\mathcal{B}^\tau\sim\mathbb{N}(0, \sigma^2_\tau)$ can be represented as a transition state between $\mathcal{B}^0$ and $\mathcal{B}^1$:
		\begin{equation}
			\mathcal{B}^\tau(i,j) \propto (\mathcal{B}^0(i,j))^{\frac{\sigma^2_0}{2\sigma_\tau^2}}(\mathcal{B}^1(i,j))^{\frac{\sigma^2_1}{2\sigma_\tau^2}}
		\end{equation}
		where $\sigma_\tau=(1-\tau)\sigma_0+\tau\sigma_1$.
	\end{proposition}
	\begin{proofA}
		Given two Gaussian blur kernels $\mathcal{B}^0\sim\mathbb{N}(0, \sigma^2_0)$ and $\mathcal{B}^1\sim\mathbb{N}(0, \sigma^2_1)$, where $\sigma_1>\sigma_0$, then in element-wise $(i,j)$,
		\begin{equation}
			\mathcal{B}^0(i,j) = \frac{1}{\sqrt{2\pi}\sigma_0}e^{-\frac{i^2+j^2}{2\sigma^2_0}},\mathcal{B}^1(i,j) = \frac{1}{\sqrt{2\pi}\sigma_1}e^{-\frac{i^2+j^2}{2\sigma^2_1}}
		\end{equation}
		given the degree of transition $\tau$, and a transitional Gaussian blur kernel $\mathcal{B}^\tau\sim\mathbb{N}(0,((1-\tau)\sigma_0+\tau\sigma_1)^2)$, then
		\begin{equation}
			\mathcal{B}^\tau(i,j) = \frac{1}{\sqrt{2\pi}((1-\tau)\sigma_0+\tau\sigma_1)}e^{-\frac{i^2+j^2}{2((1-\tau)\sigma_0+\tau\sigma_1)^2}}
		\end{equation}
		for simplifying, we transform it into logarithmic form as
		\begin{equation}
			\begin{aligned}
				log(\mathcal{B}^\tau(i,j)) &= -log(\sqrt{2\pi}((1-\tau)\sigma_0+\tau\sigma_1)) -\frac{i^2+j^2}{2((1-\tau)\sigma_0+\tau\sigma_1)^2}\\
				&=-(i^2+j^2)(\frac{1}{2\sigma^2_0}\frac{\sigma^2_0}{2((1-\tau)\sigma_0+\tau\sigma_1)^2})+\frac{1}{2\sigma^2_1}\frac{\sigma^2_1}{2((1-\tau)\sigma_0+\tau\sigma_1)^2})) + f(\tau, \sigma_0, \sigma_1)\\
				&=\frac{\sigma^2_0}{2((1-\tau)\sigma_0+\tau\sigma_1)^2}log(\mathcal{B}^0(i,j))+\frac{\sigma^2_1}{2((1-\tau)\sigma_0+\tau\sigma_1)^2}log(\mathcal{B}^1(i,j)) + f(\tau, \sigma_0, \sigma_1)
			\end{aligned}
		\end{equation}
		where $f(\tau, \sigma_0, \sigma_1)$ denotes the constant independent of $(i,j)$. Then, $\mathcal{B}^\tau$ can be represented as a transition state between $\mathcal{B}^0$ and $\mathcal{B}^1$:
		\begin{equation}
			\begin{aligned}
				\mathcal{B}^\tau(i,j) &=
				f(\tau, \sigma_0, \sigma_1)(\mathcal{B}^0(i,j))^{\frac{\sigma^2_0}{2((1-\tau)\sigma_0+\tau\sigma_1)^2}}(\mathcal{B}^1(i,j))^{\frac{\sigma^2_1}{2((1-\tau)\sigma_0+\tau\sigma_1)^2}}\\
				&\propto (\mathcal{B}^0(i,j))^{\frac{\sigma^2_0}{2\sigma_\tau^2}}(\mathcal{B}^1(i,j))^{\frac{\sigma^2_1}{2\sigma_\tau^2}}
			\end{aligned}
		\end{equation}
		where $\sigma_\tau=(1-\tau)\sigma_0+\tau\sigma_1$.
		
		\IEEEQED
	\end{proofA}
	\setcounter{equation}{\value{MYtempeqncnt}}
%

\end{document}